\documentclass[twoside, 11pt]{article}

\usepackage{jmlr2e}
\usepackage{amsmath, amstext, amsfonts, amscd, psfrag, natbib}

\newcommand{\BEAS}{\begin{eqnarray*}}
\newcommand{\EEAS}{\end{eqnarray*}}
\newcommand{\BEA}{\begin{eqnarray}}
\newcommand{\EEA}{\end{eqnarray}}
\newcommand{\BEQ}{\begin{equation}}
\newcommand{\EEQ}{\end{equation}}
\newcommand{\BIT}{\begin{itemize}}
\newcommand{\EIT}{\end{itemize}}
\newcommand{\BA}{\begin{array}{ll}}
\newcommand{\EA}{\end{array}}
\newcommand{\BP}{\begin{pmatrix}}
\newcommand{\EP}{\end{pmatrix}}
\newcommand{\BNUM}{\begin{enumerate}}
\newcommand{\ENUM}{\end{enumerate}}
\newcommand{\tr}{\hbox{trace}}
\newcommand{\diag}{\hbox{diag}}

\parskip = \baselineskip
\graphicspath{{./figures/}}

\ShortHeadings{Model Selection Through Sparse Max Likelihood Estimation}{Banerjee, El Ghaoui, and d'Aspremont}
\firstpageno{1}

\begin{document}

\title{Model Selection Through \\ Sparse Maximum Likelihood Estimation \\ for Multivariate Gaussian or Binary Data}

\author{\name Onureena Banerjee \email onureena@eecs.berkeley.edu \\
	\name Laurent El Ghaoui \email elghaoui@eecs.berkeley.edu \\
	\addr EECS Department \\	
	University of California, Berkeley \\
	Berkeley, CA 94720 USA
	\AND
	\name Alexandre d'Aspremont \email aspremon@princeton.edu \\
	\addr ORFE Department \\	
	Princeton University \\
	Princeton, NJ 08544 USA}

\editor{Leslie Pack Kaelbling}

\maketitle

\begin{abstract}%
We consider the problem of estimating the parameters of a Gaussian or binary distribution in such a way that the resulting undirected graphical model is sparse.  Our approach is to solve a maximum likelihood problem with an added $\ell_1$-norm penalty term.  The problem as formulated is convex but the memory requirements and complexity of existing interior point methods are prohibitive for problems with more than tens of nodes.  We present two new algorithms for solving problems with at least a thousand nodes in the Gaussian case.  Our first algorithm uses block coordinate descent, and can be interpreted as recursive $\ell_1$-norm penalized regression.  Our second algorithm, based on Nesterov's first order method, yields a complexity estimate with a better dependence on problem size than existing interior point methods.  Using a log determinant relaxation of the log partition function (\cite{Wain2006}), we show that these same algorithms can be used to solve an approximate sparse maximum likelihood problem for the binary case.  We test our algorithms on synthetic data, as well as on gene expression and senate voting records data.
\end{abstract}



\section{Introduction}    
%
Undirected graphical models offer a way to describe and explain the relationships among a set of variables, a central element of multivariate data analysis.  The principle of parsimony dictates that we should select the simplest graphical model that adequately explains the data.  In this paper weconsider practical ways of implementing the following approach to finding such a model:  given a set of data, we solve a maximum likelihood problem with an added $\ell_1$-norm penalty to make the resulting graph as sparse as possible.

%
Many authors have studied a variety of related ideas.  In the Gaussian case, model selection involves finding the pattern of zeros in the inverse covariance matrix, since these zeros correspond to conditional independencies among the variables.  Traditionally, a greedy forward-backward search algorithm is used to determine the zero pattern \citep[e.g.,][]{Lauritzen1996}.  However, this is computationally infeasible for data with even a moderate number of variables.  \cite{Li2005} introduce a gradient descent algorithm in which they account for the sparsity of the inverse covariance matrix by defining a loss function that is the negative of the log likelihood function.  Recently, \cite{Huan05} considered penalized maximum likelihood estimation, and \cite{Dahl05} proposed a set of large scale methods for problems where a sparsity pattern for the inverse covariance is given and one must estimate the nonzero elements of the matrix.  

Another way to estimate the graphical model is to find the set of neighbors of each node in the graph by regressing that variable against the remaining variables.  In this vein, \cite{Dobr04} employ a stochastic algorithm to manage tens of thousands of variables.  There has also been a great deal of interest in using $\ell_1$-norm penalties in statistical applications.  \cite{dasp04a} apply an $\ell_1$ norm penalty to sparse principle component analysis.  Directly related to our problem is the use of the Lasso of \cite{Tibs96} to obtain a very short list of neighbors for each node in the graph.  \cite{Mein05} study this approach in detail, and show that the resulting estimator is consistent, even for high-dimensional graphs.

%
The problem formulation for Gaussian data, therefore, is simple.  The difficulty lies in its computation.  Although the problem is convex, it is non-smooth and has an unbounded constraint set.  As we shall see, the resulting complexity for existing interior point methods is $\mathcal{O}(p^6)$, where $p$ is the number of variables in the distribution.  In addition, interior point methods require that at each step we compute and store a Hessian of size $\mathcal{O}(p^2)$.  The memory requirements and complexity are thus prohibitive for $\mathcal{O}(p)$ higher than the tens.  Specialized algorithms are needed to handle larger problems.  

%
The remainder of the paper is organized as follows.  We begin by considering Gaussian data.  In Section \ref{sec:preliminaries} we set up the problem, derive its dual, discuss properties of the solution and how heavily to weight the $\ell_1$-norm penalty in our problem.  In Section \ref{sec:block_coord_descent} we present a provably convergent block coordinate descent algorithm that can be interpreted as recursive $\ell_1$-norm penalized regression.  In Section \ref{sec:nesterov} we present a second, alternative algorithm based on Nesterov's recent work on non-smooth optimization, and give a rigorous complexity analysis with better dependence on problem size than interior point methods.  In Section \ref{sec:binary_ASML} we show that the algorithms we developed for the Gaussian case can also be used to solve an approximate sparse maximum likelihood problem for multivariate binary data, using a log determinant relaxation for the log partition function given by \cite{Wain2006}.  In Section \ref{sec:numerical_results}, we test our methods on synthetic as well as gene expression and senate voting records data.  
\section{Problem Formulation}  
\label{sec:preliminaries}

In this section we set up the sparse maximum likelihood problem for Gaussian data, derive its dual, and discuss some of its properties.

{\subsection{Problem setup.}
\label{ssec:problem_setup}}

Suppose we are given $n$ samples independently drawn from a $p$-variate Gaussian distribution:  $y^{(1)}, \ldots, y^{(n)} \sim {\mathcal N}(\Sigma_p, \mu)$, where the covariance matrix $\Sigma$ is to be estimated.  Let $S$ denote the second moment matrix about the mean:

$$
S := \frac{1}{n}\sum_{k = 1}^n (y^{(k)} - \mu)(y^{(k)} - \mu)^T.
$$

Let $\hat{\Sigma}^{-1}$ denote our estimate of the inverse covariance matrix.  Our estimator takes the form:
\BEQ
\hat{\Sigma}^{-1} = \arg \max_{X \succ 0} \log \det X - \tr(SX) - \lambda \Vert X \Vert_1.
\label{eq:SML}
\EEQ
Here, $\Vert X \Vert_1$ denotes the sum of the absolute values of the elements of the positive definite matrix $X$.  

The scalar parameter $\lambda$ controls the size of the penalty.  The penalty term is a proxy for the number of nonzero elements in $X$, and is often used -- albiet with vector, not matrix, variables -- in regression techniques, such as the Lasso.

In the case where $S \succ 0$, the classical maximum likelihood estimate is recovered for $\lambda = 0$.  However, when the number of samples $n$ is small compared to the number of variables $p$, the second moment matrix may not be invertible.  In such cases, for $\lambda > 0$, our estimator performs some regularization so that our estimate $\hat{\Sigma}$ is always invertible, no matter how small the ratio of samples to variables is.  

Even in cases where we have enough samples so that $S \succ 0$, the inverse $S^{-1}$ may not be sparse, even if there are many conditional independencies among the variables in the distribution.  By trading off maximality of the log likelihood for sparsity, we hope to find a very sparse solution that still adequately explains the data.  A larger value of $\lambda$ corresponds to a sparser solution that fits the data less well.  A smaller $\lambda$ corresponds to a solution that fits the data well but is less sparse.  The choice of $\lambda$ is therefore an important issue that will be examined in detail in Section \ref{ssec:lambda_choice}.

{\subsection{The dual problem and bounds on the solution.}
\label{ssec:dual_problem}}

We can write (\ref{eq:SML}) as 
$$
\max_{X \succ 0} \min_{\Vert U \Vert_{\infty} \leq \lambda} \log \det X + \tr(X, S + U)
$$
where $\Vert U \Vert_{\infty}$ denotes the maximum absolute value element of the symmetric matrix $U$.  This corresponds to seeking an estimate with the maximum worst-case log likelihood, over all additive perturbations of the second moment matrix $S$.  A similar robustness interpretation can be made for a number of estimation problems, such as support vector machines for classification.

We can obtain the dual problem by exchanging the max and the min.  The resulting inner problem in $X$ can be solved analytically by setting the gradient of the objective to zero and solving for $X$.  The result is
$$
\min_{\Vert U \Vert_{\infty} \leq \lambda} -\log \det (S + U) - p 
$$
where the primal and dual variables are related as: $X = (S + U)^{-1}$.  Note that the log determinant function acts a log barrier, creating an implicit constraint that $S + U \succ 0$.  

To write things neatly, let $W = S + U$.  Then the dual of our sparse maximum likelihood problem is
\BEQ
\hat{\Sigma} := \max \{ \log \det W : \Vert W - S \Vert_{\infty} \leq \lambda \}.
\label{eq:SMLdual}
\EEQ
Observe that the dual problem (\ref{eq:SML}) estimates the covariance matrix while the primal problem estimates its inverse.  We also observe that the diagonal elements of the solution are $\Sigma_{kk} = S_{kk} + \lambda$ for all $k$.

The following theorem shows that adding the $\ell_1$-norm penalty regularlizes the solution.
\begin{theorem}
For every $\lambda > 0$, the optimal solution to (\ref{eq:SML}) is unique, with bounded eigenvalues: 
$$
\frac{p}{\lambda} \geq \Vert \hat{\Sigma}^{-1} \Vert_2 \geq (\Vert S \Vert_2 + \lambda p)^{-1}. \\
$$
\label{thm:sol_bounds}
\end{theorem}
Here, $\Vert A \Vert_2$ denotes the maximum eigenvalue of a symmetric matrix $A$.

The dual problem (\ref{eq:SMLdual}) is smooth and convex.  When $p(p + 1)/2$ is in the low hundreds, the problem can be solved by existing software that uses an interior point method \citep[e.g.,][]{vandenberghe98}.  The complexity to compute an $\epsilon$-suboptimal solution using such second-order methods, however, is $\mathcal{O}(p^6\log(1/\epsilon))$, making them infeasible when $p$ is larger than the tens.  

A related problem, solved by \cite{Dahl05}, is to compute a maximum likelihood estimate of the covariance matrix when the sparsity structure of the inverse is known in advance.  This is accomplished by adding constraints to (\ref{eq:SML}) of the form: $X_{ij} = 0$ for all pairs $(i, j)$ in some specified set.  Our constraint set is unbounded as we hope to uncover the sparsity structure automatically, starting with a dense second moment matrix $S$.

{\subsection{Choice of penalty parameter.}
\label{ssec:lambda_choice}}

Consider the true, unknown graphical model for a given distribution.  This graph has $p$ nodes, and an edge between nodes $k$ and $j$ is missing if variables $k$ and $j$ are independent conditional on the rest of the variables.  For a given node $k$, let $C_k$ denote its connectivity component: the set of all nodes that are connected to node $k$ through some chain of edges.  In particular, if node $j \not \in \mathcal{C}_k$, then variables $j$ and $k$ are independent.

We would like to choose the penalty parameter $\lambda$ so that, for finite samples, the probability of error in estimating the graphical model is controlled.  To this end, we can adapt the work of \cite{Mein05} as follows.  Let $\hat{C}^{\lambda}_k$ denote our estimate of the connectivity component of node $k$.  In the context of our optimization problem, this corresponds to the entries of row $k$ in $\hat{\Sigma}$ that are nonzero. 

Let $\alpha$ be a given level in $\lbrack 0, 1 \rbrack$.  Consider the following choice for the penalty parameter in (\ref{eq:SML}):
\BEQ
\lambda(\alpha) := (\max_{i > j} \hat{\sigma}_i\hat{\sigma}_j) \frac{ t_{n - 2}(\alpha/2p^2) }{\sqrt{n - 2 + t^2_{n - 2}(\alpha/2p^2)}}
\label{eq:lambda_formula}
\EEQ

where $t_{n-2}(\alpha)$ denotes the $(100 - \alpha)$\% point of the Student's t-distribution for $n - 2$ degrees of freedom, and $\hat{\sigma}_i$ is the empirical variance of variable $i$.  Then we can prove the following theorem:
\begin{theorem}
Using $\lambda(\alpha)$ the penalty parameter in (\ref{eq:SML}), for any fixed level $\alpha$,
$$
P(\exists k \in \{1, \ldots, p\}: \hat{C}^{\lambda}_k \not \subseteq C_k) \leq \alpha.
$$
\label{thm:error_bound}
\end{theorem}

Observe that, for a fixed problem size $p$, as the number of samples $n$ increases to infinity, the penalty parameter $\lambda(\alpha)$ decreases to zero.  Thus, asymptotically we recover the classical maximum likelihood estimate, $S$, which in turn converges in probability to the true covariance $\Sigma$.  \section{Block Coordinate Descent Algorithm}
\label{sec:block_coord_descent}
In this section we present an algorithm for solving (\ref{eq:SMLdual}) that uses block coordinate descent.

{\subsection{Algorithm description.}
\label{ssec:block_coord_descent_algorithm}}

We begin by detailing the algorithm.  For any symmetric matrix $A$, let $A_{\backslash j \backslash k}$ denote the matrix produced by removing row $k$ and column $j$.  Let $A_j$ denote column $j$ with the diagonal element $A_{jj}$ removed.  The plan is to optimize over one row and column of the variable matrix $W$ at a time, and to repeatedly sweep through all columns until we achieve convergence.  

\noindent{{\bf Initialize:} $W^{(0)} := S + \lambda I$} \\
\noindent {\bf For $k \geq 0$}
\BNUM
    \item {\bf For $j = 1, \ldots, p$}

	\BNUM

		\item Let $W^{(j-1)}$ denote the current iterate.  Solve the quadratic program
\BEQ
\hat{y} := \arg \min_y \{ y^T(W_{\backslash j \backslash j}^{(j - 1)})^{-1}y : \Vert y - S_j \Vert_{\infty} \leq \lambda \}
\label{eq:qpupdate}
\EEQ

		\item Update rule: $W^{(j)}$ is $W^{(j - 1)}$ with column/row $W_j$ replaced by $\hat{y}$.

	\ENUM

	\item Let $\hat{W}^{(0)} := W^{(p)}$.  

	\item After each sweep through all columns, check the convergence condition.  Convergence occurs when
\BEQ
\tr((\hat{W}^{(0)})^{-1}S) - p + \lambda \Vert (\hat{W}^{(0)})^{-1} \Vert_1 \leq \epsilon.
\label{eq:block_coord_descent_dualitygap}
\EEQ

\ENUM

\subsection{Convergence and property of solution.}
\label{ss:convergenceproperty}

Using Schur complements, we can prove convergence:
\begin{theorem}
The block coordinate descent algorithm described above converges, acheiving an $\epsilon$-suboptimal solution to (\ref{eq:SMLdual}).  In particular, the iterates produced by the algorithm are strictly positive definite: each time we sweep through the columns, $W^{(j)} \succ 0$ for all $j$.
\label{thm:block_coord_descent_convergence}
\end{theorem}

The proof of Theorem \ref{thm:block_coord_descent_convergence} sheds some interesting light on the solution to problem (\ref{eq:SML}).  In particular, we can use this method to show that the solution has the following property:
\begin{theorem}  Fix any $k \in \{ 1, \ldots, p\}$.  If $\lambda \geq \vert S_{kj} \vert$ for all $j \ne k$, then column and row $k$ of the solution $\hat{\Sigma}$ to (\ref{eq:SMLdual}) are zero, excluding the diagonal element.
\label{thm:independence_property_SML}
\end{theorem}
This means that, for a given second moment matrix $S$, if $\lambda$ is chosen such that the condition in Theorem \ref{thm:independence_property_SML} is met for some column $k$, then the sparse maximum likelihood method estimates variable $k$ to be independent of all other variables in the distribution.  In particular, Theorem \ref{thm:independence_property_SML} implies that if $\lambda \geq \vert S_{kj} \vert$ for all $k > j$, then (\ref{eq:SML}) estimates all variables in the distribution to be pairwise independent.

Using the work of \cite{luo92}, it may be possible to show that the local convergence rate of this method is at least linear.  In practice we have found that a small number of sweeps through all columns, independent of problem size $p$, is sufficient to achieve convergence.  For a fixed number of $K$ sweeps, the cost of the method is $O(Kp^4)$, since each iteration costs $O(p^3)$.

\subsection{Interpretation as recursive penalized regression.}

The dual of (\ref{eq:qpupdate}) is
\BEQ
\min_x x^TW_{\backslash j \backslash j}^{(j - 1)}x - S_j^Tx + \lambda \Vert x \Vert_1.
\label{eq:qpduallasso}
\EEQ

Strong duality obtains so that problems (\ref{eq:qpduallasso}) and (\ref{eq:qpupdate}) are equivalent.  If we let $Q$ denote the square root of $W_{\backslash j \backslash j}^{(j - 1)}$, and $b := \frac{1}{2}Q^{-1}S_j$, then we can write (\ref{eq:qpduallasso}) as
\BEQ
\min_x \Vert Qx - b \Vert_2^2 + \lambda \Vert x \Vert_1.
\label{eq:qp_as_LASSO}
\EEQ

The problem (\ref{eq:qp_as_LASSO}) is a penalized least-squares problems, known as the Lasso.  If $W_{\backslash j \backslash j}^{(j - 1)}$ were the $j$-th principal minor of the sample covariance $S$, then (\ref{eq:qp_as_LASSO}) would be equivalent to a penalized regression of variable $j$ against all others.  Thus, the approach is reminiscent of the approach explored by \cite{Mein05}, but there are two differences.  First, we begin with some regularization and, as a consequence, each  penalized regression problem has a unique solution. Second, and more importantly, we update the problem data after each regression: except at the very first update, $W_{\backslash j \backslash j}^{(j - 1)}$ is never a minor of $S$.  In this sense, the coordinate descent method can be interpreted as a recursive Lasso.

\section{Nesterov's First Order Method}
In this section we apply the recent results due to \cite{nesterov2003} to obtain a first order algorithm for solving (\ref{eq:SML}) with lower memory requirements and a rigorous complexity estimate with a better dependence on problem size than those offered by interior point methods.  Our purpose is not to obtain another algorithm, as we have found that the block coordinate descent is fairly efficient; rather, we seek to use Nesterov's formalism to derive a rigorous complexity estimate for the problem, improved over that offered by interior-point methods.

As we will see, Nesterov's framework allows us to obtain an algorithm that has a complexity of $O(p^{4.5}/\epsilon)$, where $\epsilon>0$ is the desired accuracy on the objective of problem (\ref{eq:SML}).  This is in contrast to the complexity of interior-point methods, $O(p^6 \log (1/\epsilon))$.  Thus, Nesterov's method provides a much better dependence on problem size and lower memory requirements at the expense of a degraded dependence on accuracy.  

\subsection{Idea of Nesterov's method.}

Nesterov's method applies to a class of non-smooth, convex optimization problems of the form
\BEQ
\label{eq:gen-nest}
    \min_x \{ f(x) : x \in Q_1 \}
\EEQ
where the objective function can be written as
$$
f(x) = \hat{f}(x) + \max_u \{ \langle Ax, u \rangle_2 : u \in Q_2 \} .
$$

Here, $Q_1$ and $Q_2$ are bounded, closed, convex sets, $\hat{f}(x)$ is differentiable (with a Lipschitz-continuous gradient) and convex on $Q_1$, and $A$ is a linear operator.  The challenge is to write our problem in the appropriate form and choose associated functions and parameters in such a way as to obtain the best possible complexity estimate, by applying general results obtained by \cite{nesterov2003}.  

Observe that we can write (\ref{eq:SML}) in the form (\ref{eq:gen-nest}) if we impose bounds on the eigenvalues of the solution, $X$.   To this end, we let
\BEQ
\BA
Q_1 := \{ x: a I \preceq X \preceq b I \} \\
\\
Q_2 := \{ u : \Vert u \Vert_{\infty} \leq \lambda \} \\
\EA
\EEQ
where the constants $a, b$ are given such that $b > a > 0$.   By Theorem \ref{thm:sol_bounds}, we know that such bounds always exist.  We also define $\hat{f}(x) := -\log \det x + \langle S, x \rangle$, and $A := \lambda I$.  

To $Q_1$ and $Q_2$, we associate norms and continuous, strongly convex functions, called prox-functions, $d_1(x)$ and $d_2(u)$.  For $Q_1$ we choose the Frobenius norm, and a prox-function $d_1(x) = -\log \det x + \log b$. For $Q_2$, we choose the Frobenius norm again, and a prox-function $d_2(x) = \Vert u \Vert_F^2/2$. 

The method applies a smoothing technique to the non-smooth problem (\ref{eq:gen-nest}), which replaces the objective of the original problem, $f(x)$, by a penalized function involving the prox-function $d_2(u)$: 
\BEQ
\label{eq:feps-def}
    \tilde{f}(x) = \hat{f}(x) + \max_{u \in Q_2} \{\langle Ax,u \rangle - \mu d_2(u) \}.
\EEQ

The above function turns out to be a smooth uniform approximation to $f$ everywhere.  It is differentiable, convex on $Q_1$, and a has a Lipschitz-continuous gradient, with a constant $L$ that can be computed as detailed below.  A specific gradient scheme is then applied to this smooth approximation, with convergence rate $O(L/\epsilon)$. 

\subsection{Algorithm and complexity estimate.}

To detail the algorithm and compute the complexity, we must first calculate some parameters corresponding to our definitions and choices above.  First, the strong convexity parameter for $d_1(x)$ on $Q_1$ is $\sigma_1 = 1/b^2$, in the sense that 

$$\nabla^2 d_1(X)[H,H] = \hbox{trace}(X^{-1}H X^{-1}H) \ge b^{-2}\|H\|_F^2 $$

for every symmetric $H$. Furthermore, the center of the set $Q_1$ is $x_0:=\arg\min_{x \in Q_1} d_1(x) \ = b I$,  and satisfies $d_1(x_0) = 0$. With our choice, we have $D_1 := \max_{x \in Q_1} \: d_1(x) = p\log (b/a)$.

Similarly, the strong convexity parameter for $d_2(u)$ on $Q_2$ is $\sigma_2 := 1$, and we have 

$$D_2 := \max_{u \in Q_2} d_2(U) = p^2/2.$$

With this choice, the center of the set $Q_2$ is $u_0 := \arg\min_{u \in Q_2} d_2(u) = 0$.  

For a desired accuracy $\epsilon$, we set the smoothness parameter $\mu := \epsilon/2D_2$, and set $x_0 = b I$. The algorithm proceeds as follows:

\noindent {\bf For $k \geq 0$ do}
\begin{enumerate}

    \item Compute $\nabla \tilde{f}(x_k) = -x^{-1} + S + u^\ast(x_k)$, where $u^\ast(x)$
    solves (\ref{eq:feps-def}).

    \item Find $y_k = \arg\min_{y} \: \{\langle \nabla \tilde{f}(x_k) , y-x_k \rangle +
\frac{1}{2} L(\epsilon) \|y-x_k\|_F^2 ~:~ y  \in Q_1 \}$.

    \item Find $z_k = \arg\min_{x} \: \{ \frac{L(\epsilon)}{\sigma_1} d_1(X) +
\sum_{i=0}^k \frac{i+1}{2} \langle \nabla \tilde{f}(x_i),x-x_i \rangle  ~:~ x  \in Q_1 \}$.

    \item Update $x_k = \frac{2}{k+3} z_k + \frac{k+1}{k+3} y_k$.

\end{enumerate}

In our case, the Lipschitz constant for the gradient of our smooth approximation to the objective function is 

$$L(\epsilon) := M+D_2\|A\|^2/(2 \sigma_2\epsilon)$$

where $M := 1/a^2$ is the Lipschitz constant for the gradient of $\tilde{f}$, and the norm $\Vert A \Vert$ is induced by the Frobenius norm, and is equal to $\lambda$.  

The algorithm is guaranteed to produce an $\epsilon$-suboptimal solution after a number of steps not exceeding
\BEQ
\label{eq:N-def}
\BA 
N(\epsilon) := \\
\\
4 \|A\| \sqrt{\displaystyle\frac{D_1D_2}{\sigma_1\sigma_2} } \cdot
\frac{1}{\epsilon} + \sqrt{\frac{MD_1}{\sigma_1\epsilon}} \\
\\
= (\kappa \sqrt{(\log \kappa)})( 4 p^{1.5} a \lambda/\sqrt{2} + \sqrt{\epsilon p})/\epsilon.
\EA
\EEQ
where $\kappa = b/a$ is a bound on the condition number of the solution. 

Now we are ready to estimate the complexity of the algorithm.  For Step $1$, the gradient of the smooth approximation is computed in closed form by taking the inverse of $x$.  Step $2$ essentially amounts to projecting on $Q_1$, and requires that we solve an eigenvalue problem.  The same is true for Step $3$.  In fact, each iteration costs $O(p^3)$. The number of iterations necessary to achieve an objective with absolute accuracy less than $\epsilon$ is given in (\ref{eq:N-def}) by $N(\epsilon) = O(p^{1.5}/\epsilon)$.  Thus, if the condition number $\kappa$ is fixed in advance, the complexity of the algorithm is $O(p^{4.5}/\epsilon)$.

\section{Binary Variables: Approximate Sparse Maximum Likelihood Estimation}
In this section, we consider the problem of estimating an undirected graphical model for multivariate binary data.  Recently, \cite{Wain06_2} applied an $\ell_1$-norm penalty to the logistic regression problem to obtain a binary version of the high-dimensional consistency results of \cite{Mein05}.  We apply the log determinant relaxation of \cite{Wain2006} to formulate an approximate sparse maximum likelihood (ASML) problem for estimating the parameters in a multivariate binary distribution.   We show that the resulting problem is the same as the Gaussian sparse maximum likelihood (SML) problem, and that we can therefore apply our previously-developed algorithms to sparse model selection in a binary setting.  

Consider a distribution made up of $p$ binary random variables.  Using $n$ data samples, we wish to estimate the structure of the distribution.  The logistic model of this distribution is
\BEQ
p(x; \theta) = \exp \{ \sum_{i = 1}^p \theta_i x_i + \sum_{i = 1}^{p - 1} \sum_{j = i + 1}^p \theta_{ij}x_ix_j - A(\theta) \}
\label{eq:binary_density}
\EEQ
where 
\BEQ
A(\theta) = \log \sum_{x \in {\mathcal X}^p} \exp \{ \sum_{i = 1}^p \theta_i x_i + \sum_{i = 1}^{p - 1} \sum_{j = i + 1}^p \theta_{ij}x_ix_j  \}
\EEQ
is the log partition function.  

The sparse maximum likelihood problem in this case is to maximize (\ref{eq:binary_density}) with an added $\ell_1$-norm penalty on terms $\theta_{kj}$.  Specifically, in the undirected graphical model, an edge between nodes $k$ and $j$ is missing if $\theta_{kj} = 0$.

A well-known difficulty is that the log partition function has too many terms in its outer sum to compute.  However, if we use the log determinant relaxation for the log partition function developed by \cite{Wain2006}, we can obtain an approximate sparse maximum likelihood (ASML) estimate.  We shall set up the problem in the next section.

{\subsection{Problem formulation.}
\label{ssec:binary_problem_formulation}}
Let's begin with some notation.  Letting $d := p(p+1)/2$, define the map $R : {\bf R}^d \rightarrow S^{p + 1}$ as follows:

$$
R(\theta) = 
\BP
0 & \theta_1 & \theta_2 & \ldots & \theta_p \\
\theta_1 & 0 & \theta_{12} & \ldots & \theta_{1p} \\
\vdots \\
\theta_p & \theta_{1p} & \theta_{2p} & \ldots & 0 \\
\EP
$$

Suppose that our $n$ samples are $z^{(1}), \ldots, z^{(n)} \in \{ -1, +1 \}^p$.  Let $\bar{z}_i$ and $\bar{z}_{ij}$ denote sample mean and second moments.  The sparse maximum likelihood problem is
\BEQ
\hat{\theta}_{\hbox{exact}} := \arg \max_{\theta} \frac{1}{2}\langle R(\theta), R(\bar{z}) \rangle - A(\theta) - \lambda \Vert \theta \Vert_1.
\label{eq:binarySML}
\EEQ

Finally define the constant vector $m = (1, \frac{4}{3}, \ldots, \frac{4}{3}) \in {\bf R}^{p + 1}$.  \cite{Wain2006} give an upper bound on the log partition function as the solution to the following variational problem:
\BEQ
\BA
A(\theta) \leq \max_{\mu} \frac{1}{2}\log \det (R(\mu) + \diag(m)) + \langle \theta, \mu \rangle \\
\\
= \frac{1}{2} \cdot \max_{\mu} \log \det (R(\mu) + \diag(m)) + \langle R(\theta), R(\mu) \rangle. \\
\EA
\label{eq:wainwright_bound}
\EEQ

If we use the bound (\ref{eq:wainwright_bound}) in our sparse maximum likelihood problem (\ref{eq:binarySML}), we won't be able to extract an optimizing argument $\hat{\theta}$.  Our first step, therefore, will be to rewrite the bound in a form that will allow this.

\begin{lemma}
We can rewrite the bound (\ref{eq:wainwright_bound}) as
\BEQ
A(\theta) \leq \frac{p}{2}\log(\frac{e\pi}{2}) - \frac{1}{2}(p + 1) - \frac{1}{2} \cdot \{ \max_{\nu} \nu^Tm + \log\det (- (R(\theta) + \hbox{diag}(\nu)) ).
\label{eq:modified_partition_bound}
\EEQ
\label{lem:log_partition_bound}
\end{lemma}

Using this version of the bound (\ref{eq:wainwright_bound}), we have the following theorem.

\begin{theorem}
Using the upper bound on the log partition function given in (\ref{eq:modified_partition_bound}), the approximate sparse maximum likelihood problem has the following solution: 
\BEQ
\BA
\hat{\theta}_k = \bar{\mu}_k  \\
\\ 
\hat{\theta}_{kj} = -(\hat{\Gamma})^{-1}_{kj} \\
\EA
\label{eq:binary_ASML_solution}
\EEQ
where the matrix $\hat{\Gamma}$ is the solution to the following problem, related to (\ref{eq:SMLdual}):
\BEQ
\hat{\Gamma} := \arg \max \{ \log\det W : W_{kk} = S_{kk} + \frac{1}{3}, \hbox{    } \vert W_{kj} - S_{kj} \vert \leq \lambda \}.
\label{eq:binaryASML_formulation}
\EEQ
\label{thm:connect_ASML_SML}
\end{theorem} 

Here, $S$ is defined as before:
$$
S = \frac{1}{n}\sum_{k = 1}^n (z^{(k)} - \bar{\mu})(z^{(k)} - \bar{\mu})^T
$$
where $\bar{\mu}$ is the vector of sample means $\bar{z}_i$.  

In particular, this means that we can reuse the algorithms developed in Sections \ref{sec:block_coord_descent} and \ref{sec:nesterov} for problems with binary variables.   The relaxation (\ref{eq:wainwright_bound}) is the simplest one offered by \cite{Wain2006}.  The relaxation can be tightened by adding linear constraints on the variable $\mu$.

{\subsection{Penalty parameter choice for binary variables.}
\label{ssec:binary_lambda_choice}}

For the choice of the penalty parameter $\lambda$, we can derive a formula analogous to (\ref{eq:lambda_formula}).  Consider the choice
\BEQ
\lambda(\alpha)_{\hbox{bin}} := \frac{(\chi^2(\alpha/2p^2, 1))^{\frac{1}{2}} }{(\min_{i > j} \hat{\sigma}_i\hat{\sigma}_j)\sqrt{n}}  
\label{eq:binary_lambda_formula}
\EEQ

where $\chi^2(\alpha, 1)$ is the $(100 - \alpha)$\% point of the chi-square distribution for one degree of freedom.  Since our variables take on values in $\{-1, 1\}$, the empirical variances are of the form:
$$
\hat{\sigma}_i^2 = 1 - \bar{\mu}_i^2.
$$

Using (\ref{eq:binary_lambda_formula}), we have the following binary version of Theorem \ref{thm:error_bound}:
\begin{theorem}
With (\ref{eq:binary_lambda_formula}) chosen as the penalty parameter in the approximate sparse maximum likelihood problem, for a fixed level $\alpha$,
$$
P(\exists k \in \{1, \ldots, p\}: \hat{C}^{\lambda}_k \not \subseteq C_k) \leq \alpha.
$$
\label{thm:error_bound_binary}
\end{theorem}

\section{Numerical Results}
\label{sec:numerical_results}
In this section we present the results of some numerical experiments, both on synthetic and real data.  

\subsection{Synthetic experiments.}

Synthetic experiments require that we generate underlying sparse inverse covariance matrices.  To this end, we first randomly choose a diagonal matrix with positive diagonal entries.  A given number of nonzeros are inserted in the matrix at random locations symmetrically.  Positive definiteness is ensured by adding a multiple of the identity to the matrix if needed.  The multiple is chosen to be only as large as necessary for inversion with no errors.

\subsubsection{Sparsity and thresholding.}

\begin{figure}[ht]
\vskip 0.2in
\begin{center}
\setlength{\epsfxsize}{4in}
\centerline{\epsfbox{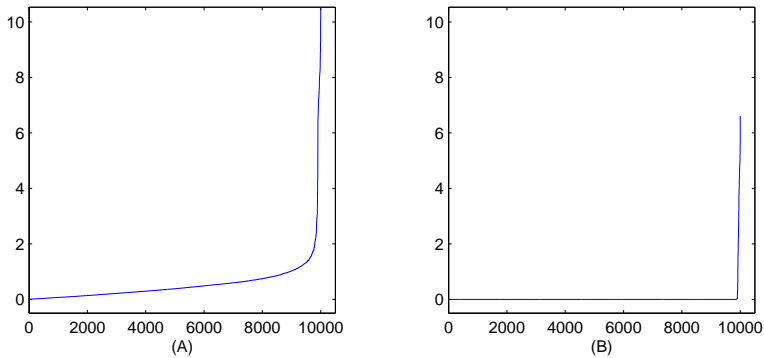}}
\vskip 0.1in
\caption{Sorted absolute value of elements of (A) $S^{-1}$ and (B) $\hat{\Sigma}^{-1}$.  The solution $\hat{\Sigma}^{-1}$ to (\ref{eq:SML}) is un-thresholded.}
\label{fig:sparsitysorted}
\end{center}
\vskip -0.2in
\end{figure}

A very simple approach to obtaining a sparse estimate of the inverse covariance matrix would be to apply a threshold to the inverse empirical covariance matrix, $S^{-1}$.  However, even when $S$ is easily invertible, it can be difficult to select a threshold level.  We solved a synthetic problem of size $p = 100$ where the true concentration matrix density was set to $\delta = 0.1$.  Drawing $n = 200$ samples, we plot in Figure (\ref{fig:sparsitysorted}) the sorted absolute value elements of $S^{-1}$ on the left and $\hat{\Sigma}^{-1}$ on the right.  

It is clearly easier to choose a threshold level for the  SML estimate.  Applying a threshold to either $S^{-1}$ or $\hat{\Sigma}^{-1}$ would decrease the log likelihood of the estimate by an unknown amount.  We only observe that to preserve positive definiteness, the threshold level $t$ must satisfy the bound
$$
t \leq \min_{\Vert v \Vert_1 \leq 1} v^TS^{-1}v.
$$

\subsubsection{Recovering structure.} 

We begin with a small experiment to test the ability of the method to recover the sparse structure of an underlying covariance matrix.  Figure \ref{fig:30pattern} (A) shows a sparse inverse covariance matrix of size $p = 30$.  Figure \ref{fig:30pattern} (B) displays a corresponding $S^{-1}$, using $n = 60$ samples.  Figure {\ref{fig:30pattern} (C) displays the solution to (\ref{eq:SML}) for $\lambda = 0.1$.  The value of the penalty parameter here is chosen arbitrarily, and the solution is not thresholded.  Nevertheless, we can still pick out features that were present in the true underlying inverse covariance matrix.

\begin{figure}[ht]
\vskip 0.2in
\begin{center}
\setlength{\epsfxsize}{4.5in}
\centerline{\epsfbox{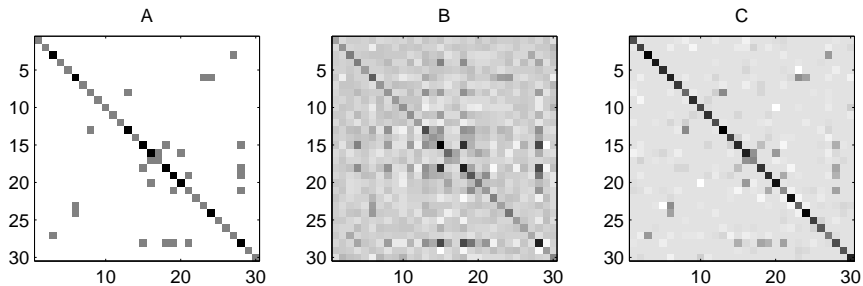}}
\caption{Recovering the sparsity pattern. We plot (A) the original inverse covariance matrix 
    $\Sigma^{-1}$, (B) the noisy sample inverse $S^{-1}$, and (C) the solution to problem (\ref{eq:SML}) 
for $\lambda=0.1$.}
\label{fig:30pattern}
\end{center}
\vskip -0.2in
\end{figure} 

Using the same underlying inverse covariance matrix, we repeat the experiment using smaller sample sizes.  We solve (\ref{eq:SML}) for $n = 30$ and $n = 20$ using the same arbitrarily chosen penalty parameter value $\lambda = 0.1$, and display the solutions in Figure (\ref{fig:patternvariousn}).  As expected, our ability to pick out features of the true inverse covariance matrix diminishes with the number of samples.  This is an added reason to choose a larger value of $\lambda$ when we have fewer samples, as in (\ref{eq:lambda_formula}).

\begin{figure}[ht]
\vskip 0.2in
\begin{center}
\setlength{\epsfxsize}{4.5in}
\centerline{\epsfbox{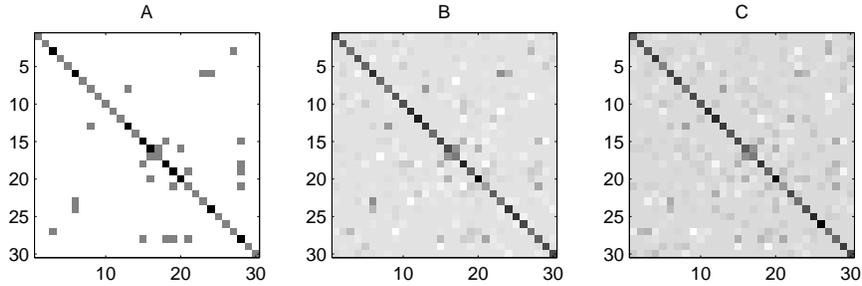}}
\caption{Recovering the sparsity pattern for small sample size. We plot (A) the original inverse covariance matrix 
    $\Sigma^{-1}$, (B) the solution to problem (\ref{eq:SML}) for $n = 30$ and (C) the solution for
    $n = 20$. A penalty parameter of $\lambda=0.1$ is used for (B) and (C).}
\label{fig:patternvariousn}
\end{center}
\vskip -0.2in
\end{figure}

\subsubsection{Path following experiments.}

Figure (\ref{fig:pathfollowing}) shows two path following examples.  We solve two randomly generated problems of size $p = 5$ and $n = 100$ samples.  The red lines correspond to elements of the solution that are zero in the true underlying inverse covariance matrix.  The blue lines correspond to true nonzeros.  The vertical lines mark ranges of $\lambda$ for which we recover the correct sparsity pattern exactly.  Note that, by Theorem \ref{thm:independence_property_SML}, for $\lambda$ values greater than those shown, the solution will be diagonal.

\begin{figure*}[t]
\hbox{\hskip -0.55in
\setlength{\epsfxsize}{3in}
\psfrag{Elements of solution}{$\hat{\Sigma}^{-1}_{ij}$}
\psfrag{lambda}{$\lambda$}
\epsfbox{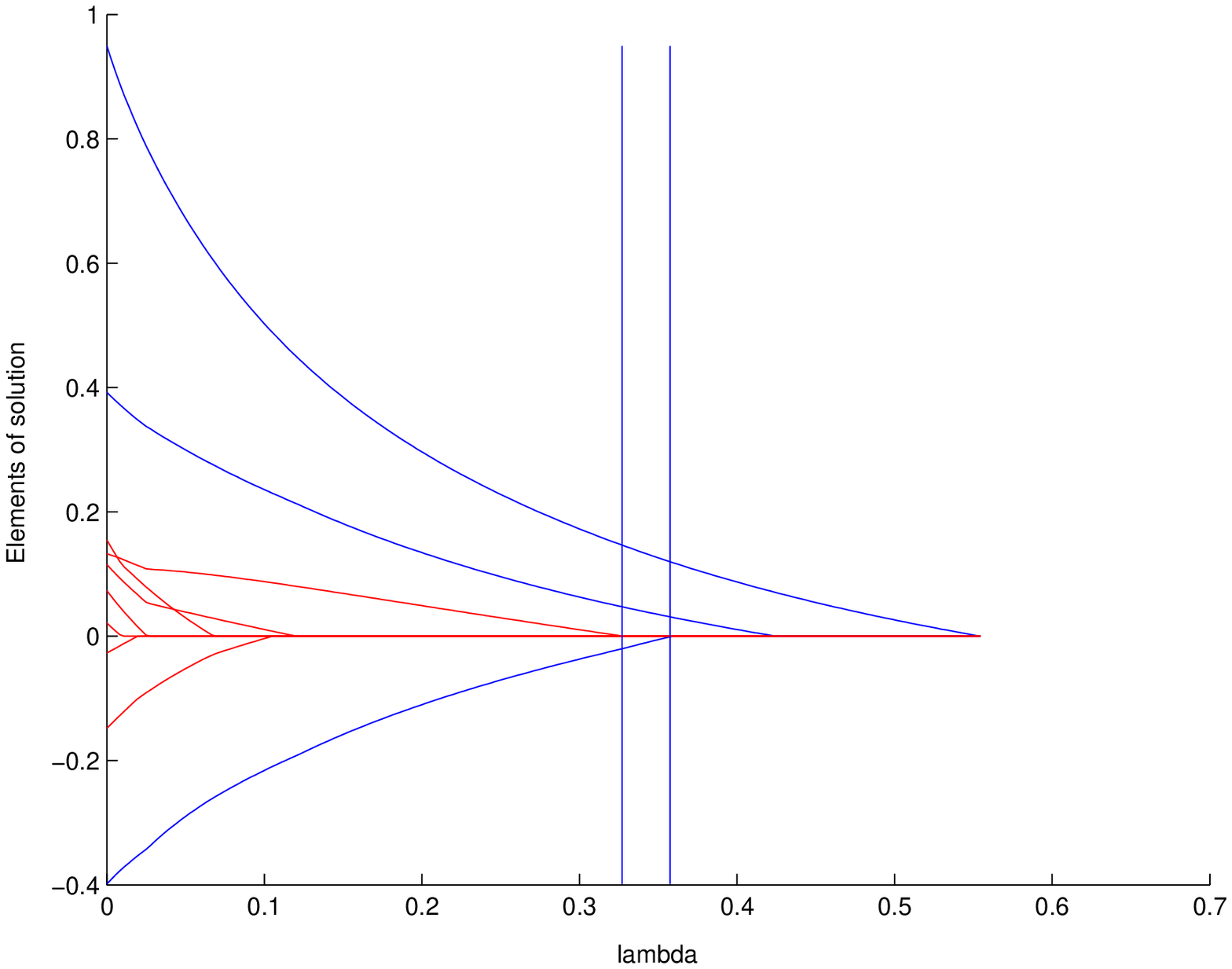}
\hskip 0.2in
\setlength{\epsfxsize}{3in}
\psfrag{lambda}{$\lambda$}
\psfrag{Elements of solution}{$\hat{\Sigma}^{-1}_{ij}$}
\epsfbox{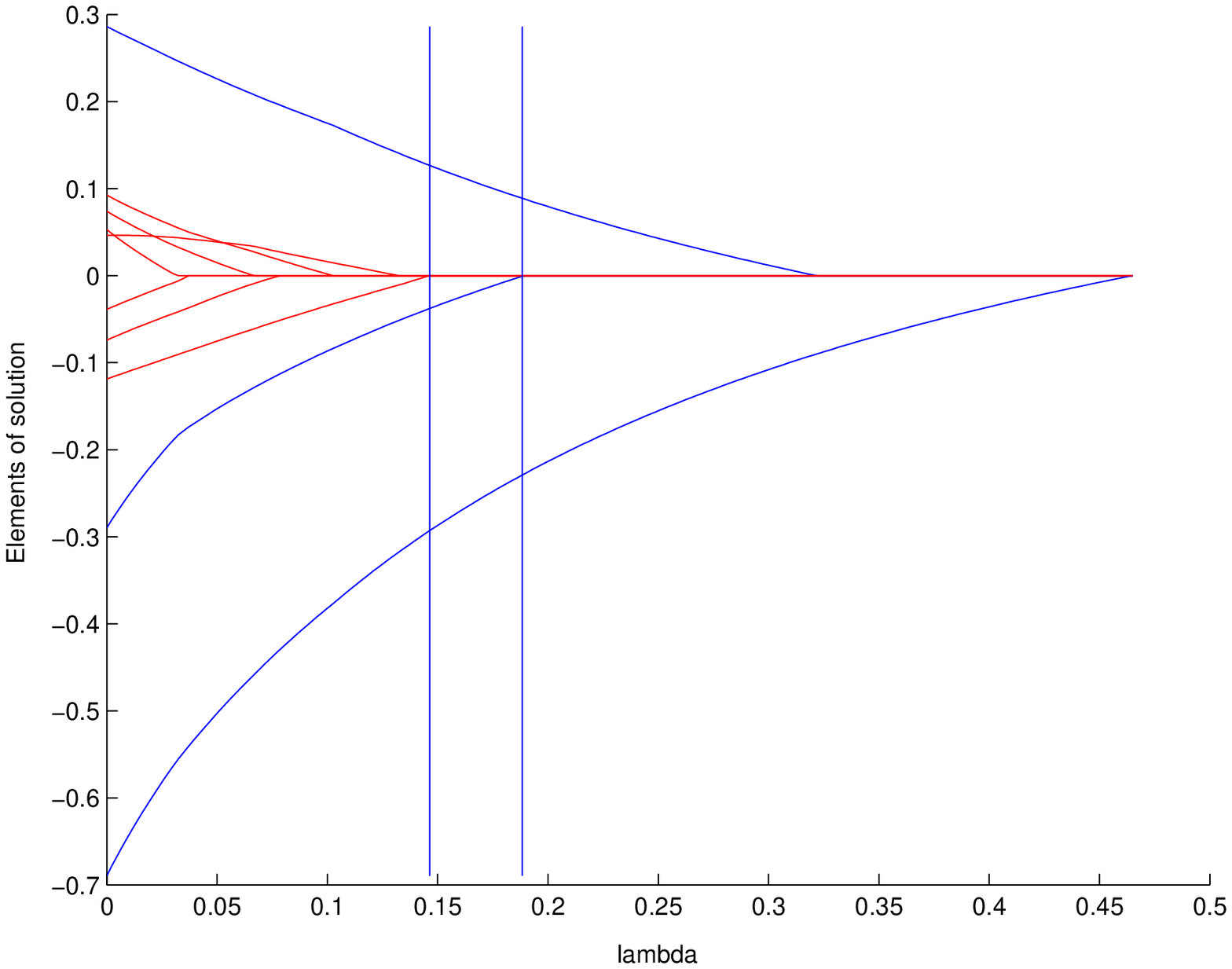}
}
\vfill
\caption{Path following: elements of solution to (\ref{eq:SML}) as $\lambda$ increases.  Red lines correspond to elements that are zero in the true inverse covariance matrix; blue lines correspond to true nonzeros.  Vertical lines mark a range of $\lambda$ values using which we recover the sparsity pattern exactly.}
\label{fig:pathfollowing}
\end{figure*}

On a related note, we observe that (\ref{eq:SML}) also works well in recovering the sparsity pattern of a matrix masked by noise.  The following experiment illustrates this observation.  We generate a sparse inverse covariance matrix of size $p = 50$ as described above.  Then, instead of using an empirical covariance $S$ as input to (\ref{eq:SML}), we use $S = (\Sigma^{-1} + V)^{-1}$, where $V$ is a randomly generated uniform noise of size $\sigma = 0.1$.  We then solve (\ref{eq:SML}) for various values of the penalty parameter $\lambda$.


In figure \ref{fig:maskingexperiment}, for a each of value of $\lambda$ shown, we randomly selected 10 sample covariance matrices $S$ of size $p=50$ and computed the number of misclassified zeros and nonzero elements in the solution to (\ref{eq:SML}).  We plot the average percentage of errors (number of misclassified zeros plus misclassified nonzeros divided by $p^2$), as well as error bars corresponding to one standard deviation.  As shown, the error rate is nearly zero on average when the penalty is set to equal the noise level $\sigma$.


\begin{figure}[t]
\setlength{\epsfxsize}{4in}
\psfrag{err}{Error (in \%)}
\psfrag{rhob}{$\log(\lambda/\sigma)$}
\epsfbox{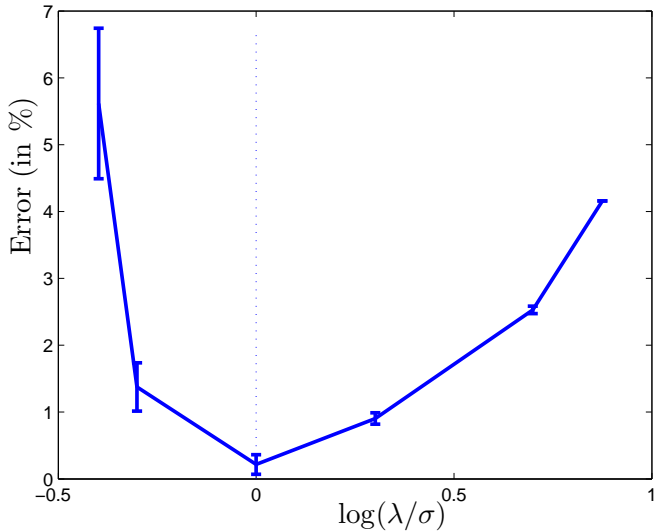}
\vskip 0.1in
\caption{Recovering sparsity pattern in a matrix with added uniform noise of size $\sigma = 0.1$.  We plot the average percentage or misclassified entries as a function of $\log(\lambda/\sigma)$.}
\label{fig:maskingexperiment} 
\end{figure}

\subsubsection{CPU times versus problem size.}

For a sense of the practical performance of the Nesterov method and the block coordinate descent method, we randomly selected 10 sample covariance matrices $S$ for problem sizes $p$ ranging from $400$ to $1000$.In each case, the number of samples $n$ was chosen to be about a third of $p$.  In figure \ref{fig:1000cputimes} we plot the average CPU time to achieve a duality gap of $\epsilon = 1$.   CPU times were computed using an AMD Athlon 64 2.20Ghz processor with 1.96GB of RAM.

\begin{figure}[ht]
\vskip 0.2in
\begin{center}
\setlength{\epsfxsize}{3.5in}
\centerline{\epsfbox{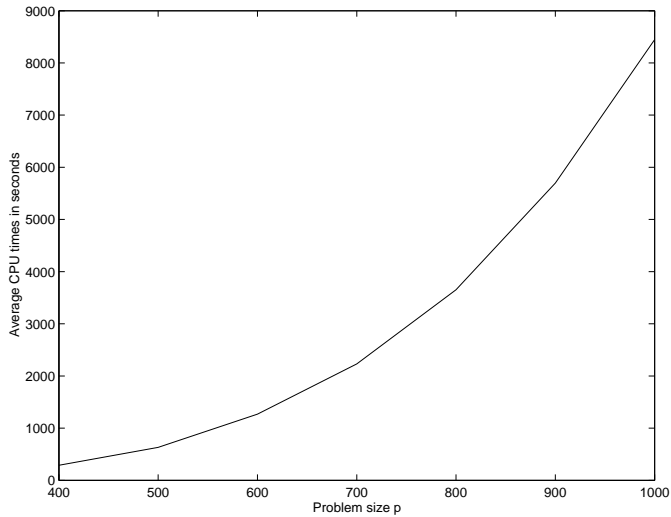}}
\caption{Average CPU times vs. problem size using block coordinate descent.  We plot the average CPU time (in seconds)
to reach a gap of $\epsilon = 0.1$ versus problem size $p$.}
\label{fig:1000cputimes}
\end{center}
\vskip -0.2in
\end{figure} 

As shown, we are typically able to solve a problem of size $p = 1000$ in about two and half hours.

\subsubsection{Performance as a binary classifier.}

In this section we numerically examine the ability of the sparse maximum likelihood (SML) method to correctly classify elements of the inverse covariance matrix as zero or nonzero.  For comparision, we will use the Lasso estimate of \cite{Mein05}, which has been shown to perform extremely well.  The Lasso regresses each variable against all others one at a time.  Upon obtaining a solution $\theta^{(k)}$ for each variable $k$, one can estimate sparsity in one of two ways: either by declaring an element $\hat{\Sigma}_{ij}$ nonzero if both $\theta^{(k)}_i \ne 0$ and $\theta^{(k)}_j \ne 0$ (Lasso-AND) or, less conservatively, if either of those quantities is nonzero (Lasso-OR).

As noted previously, \cite{Mein05} have also derived a formula for choosing their penalty parameter.  Both the SML and Lasso penalty parameter formulas depend on a chosen level $\alpha$, which is a bound on the same error probability for each method.  For these experiments, we set $\alpha = 0.05$.

In the following experiments, we fixed the problem size $p$ at 30 and generated sparse underlying inverse covariance matrices as described above.  We varied the number of samples $n$ from 10 to 310.  For each value of $n$ shown, we ran 30 trials in which we estimated the sparsity pattern of the inverse covariance matrix using the SML, Lasso-OR, and Lasso-AND methods.  We then recorded the average number of nonzeros estimated by each method, and the average number of entries correctly identified as nonzero (true positives).

We show two sets of plots.  Figure (\ref{fig:lowdensity}) corresponds to experiments where the true density was set to be low, $\delta = 0.05$.  We plot the power (proportion of correctly identified nonzeros), positive predictive value (proportion of estimated nonzeros that are correct), and the density estimated by each method.  Figure (\ref{fig:highdensity}) corresponds to experiments where the true density was set to be high, $\delta = 0.40$, and we plot the same three quantities.  

\begin{figure*}[t]
\hbox{
\hskip -0.6in
\setlength{\epsfxsize}{3.5in}
\epsfbox{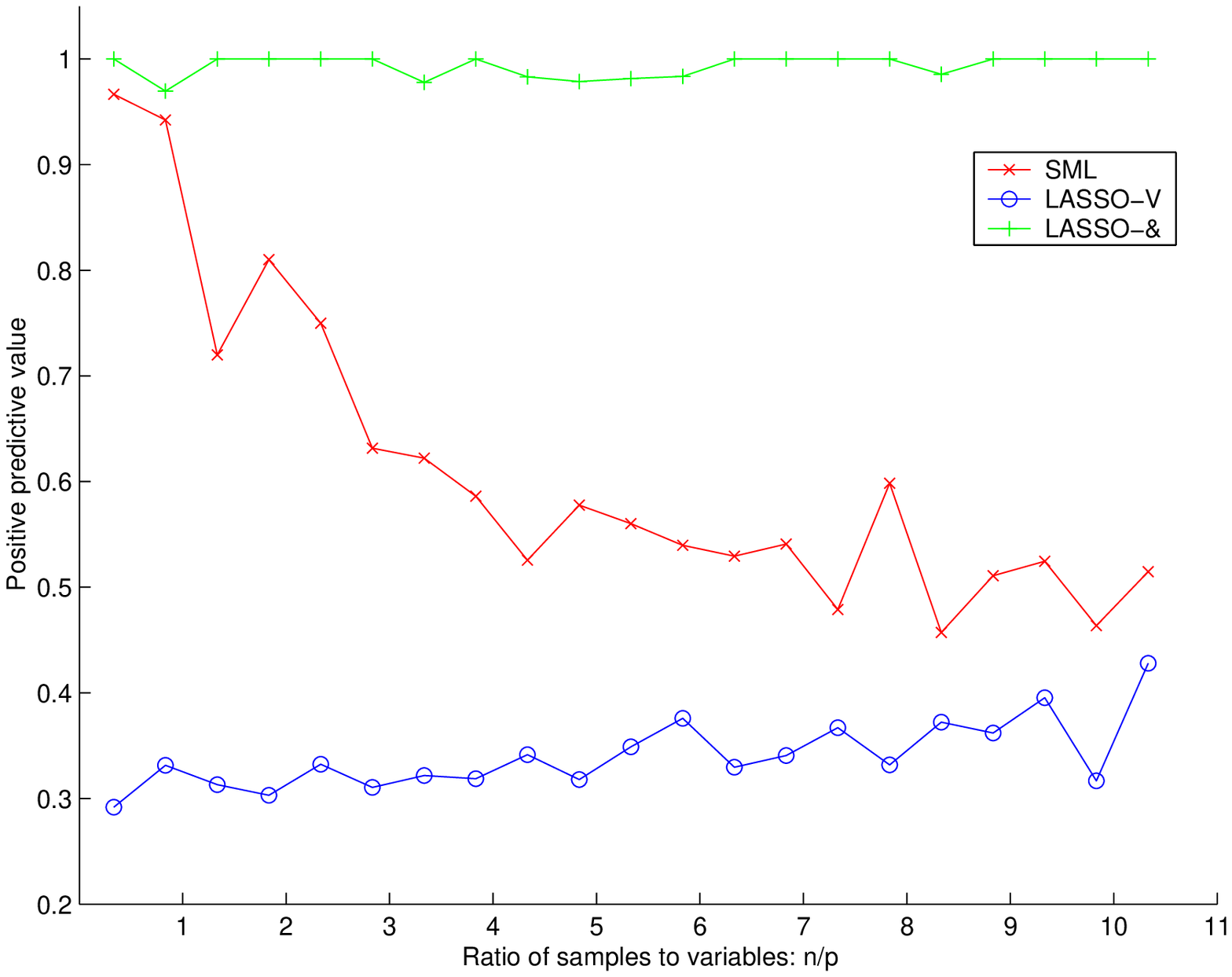}
\setlength{\epsfxsize}{3.5in}
\epsfbox{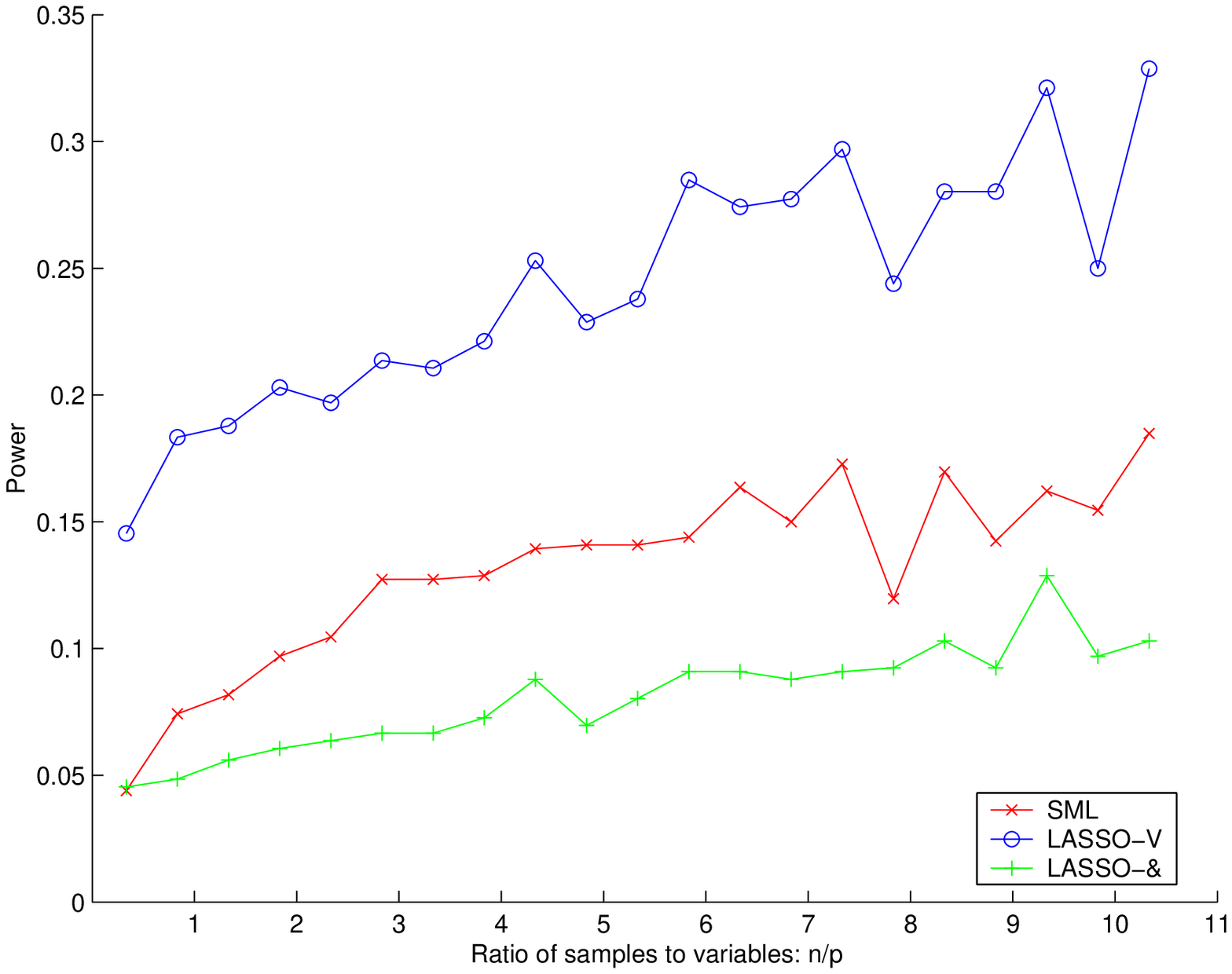}
}
\vfill
\setlength{\epsfxsize}{3.5in}
\centerline{\epsfbox{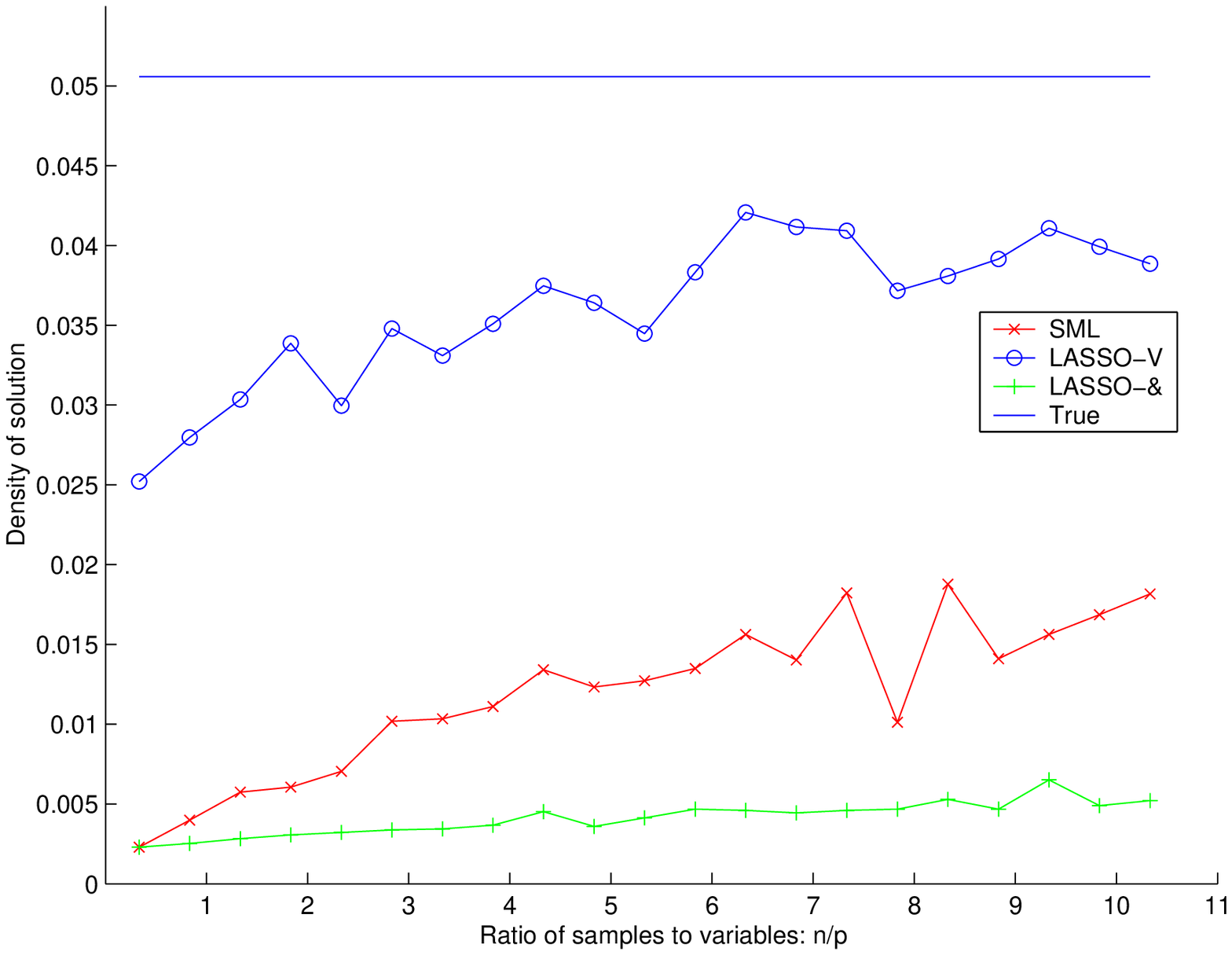}}
\caption{Classifying zeros and nonzeros for a true density of $\delta = 0.05$.  We plot the positive predictive value, the power, and the estimated density using SML, Lasso-OR and Lasso-AND.}
\label{fig:lowdensity}
\end{figure*}

\begin{figure*}[t]
\hbox{
\hskip -0.6in
\setlength{\epsfxsize}{3.5in}
\epsfbox{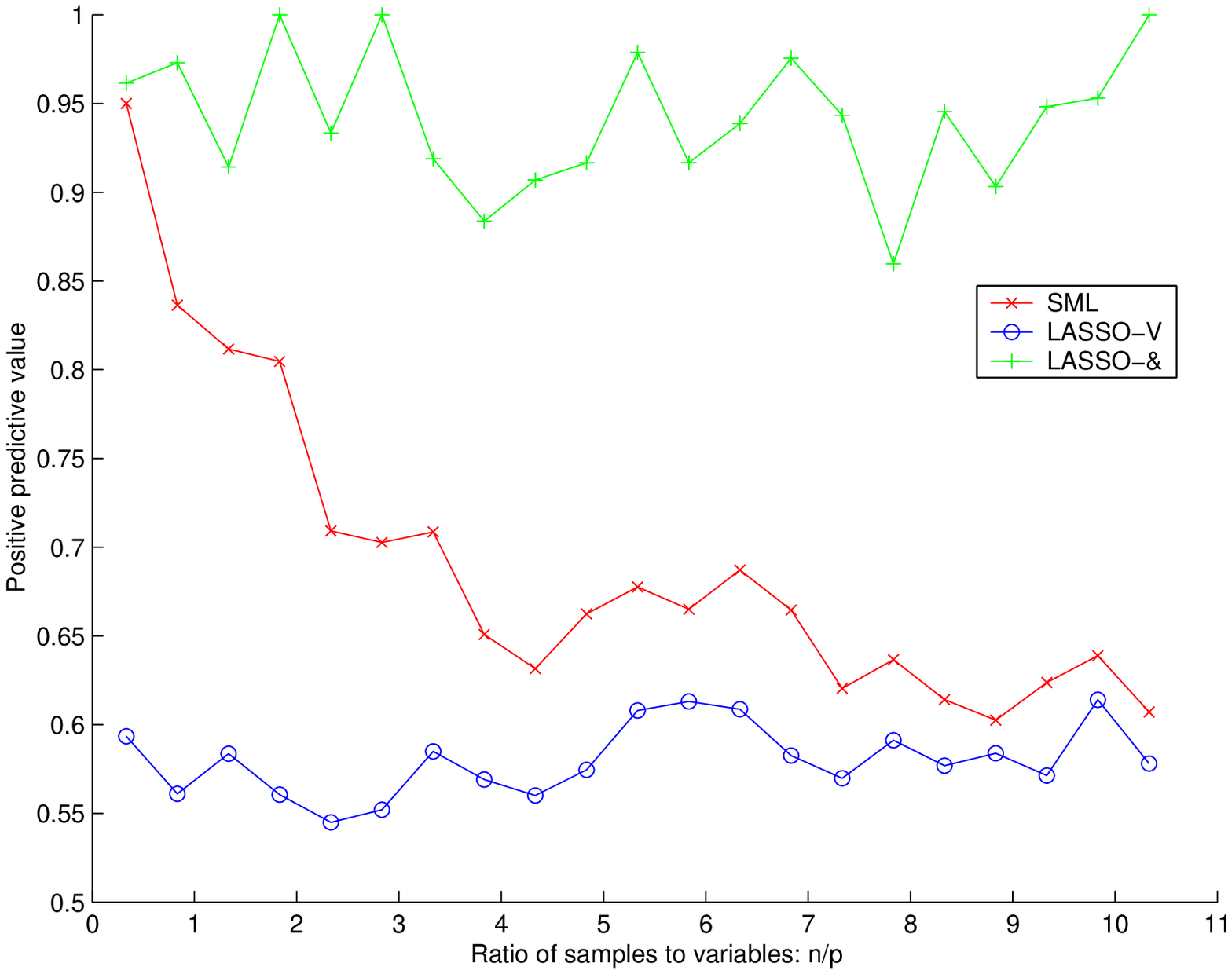}
\setlength{\epsfxsize}{3.5in}
\epsfbox{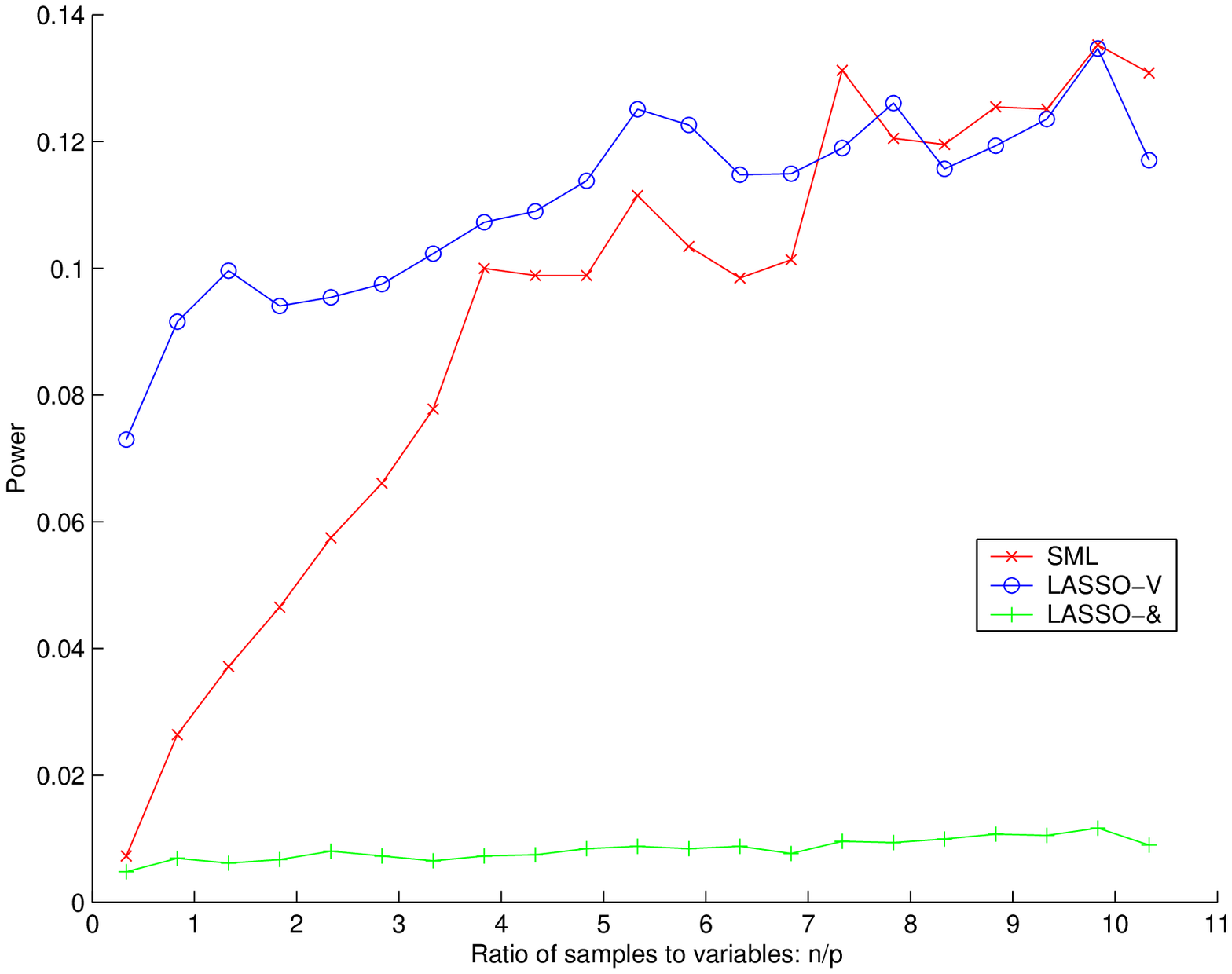}
}
\vfill
\setlength{\epsfxsize}{3.5in}
\centerline{\epsfbox{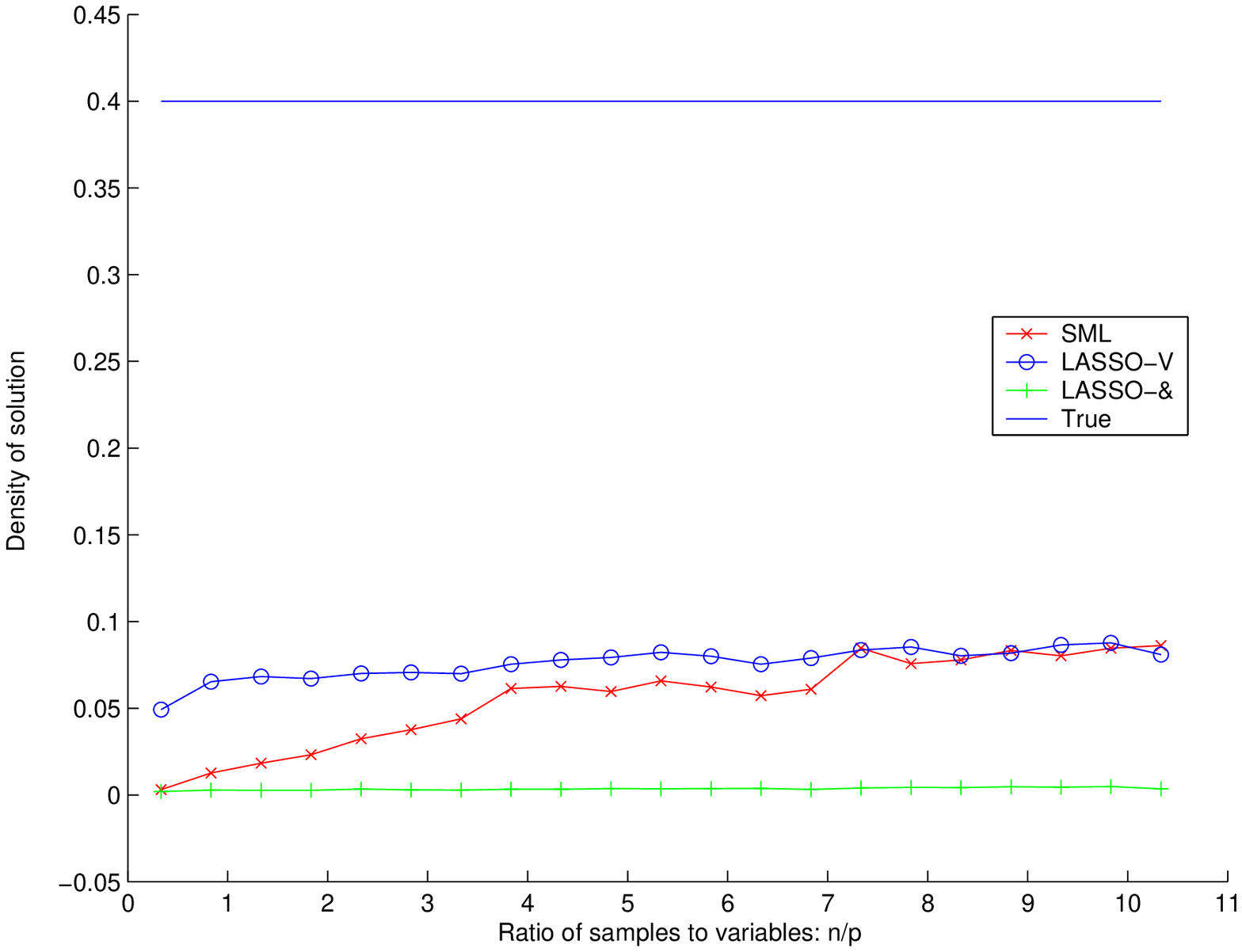}}
\caption{Classifying zeros and nonzeros for a true density of $\delta = 0.40$.  We plot the positive predictive value, the power, and the estimated density using SML, Lasso-OR and Lasso-AND.}
\label{fig:highdensity}
\end{figure*}

\cite{Mein05} report that, asymptotically, Lasso-AND and Lasso-OR yield the same estimate of the sparsity pattern of the inverse covariance matrix.  At a finite number of samples, the SML method seems to fall in in between the two methods in terms of power, positive predictive value, and the density of the estimate.  It typically offers, on average, the lowest total number of errors, tied with either Lasso-AND or Lasso-OR.  Among the two Lasso methods, it would seem that if the true density is very low, it is slightly better to use the more conservative Lasso-AND.  If the density is higher, it may be better to use Lasso-OR.  When the true density is unknown, we can achieve an accuracy comparable to the better choice among the Lasso methods by computing the SML estimate.  Figure (\ref{fig:recover_vs_lasso}) shows one example of sparsity pattern recovery when the true density is low.

\begin{figure}[ht]
\vskip 0.2in
\begin{center}
\setlength{\epsfxsize}{4.5in}
\centerline{\epsfbox{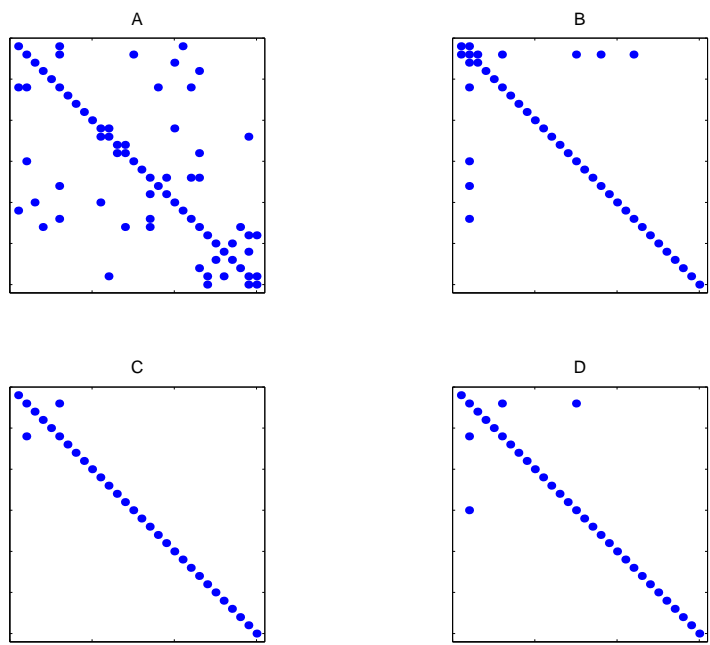}}
\caption{Comparing sparsity pattern recovery to the Lasso.  (A) true covariance (B) Lasso-OR (C) Lasso-AND (D) SML.}
\label{fig:recover_vs_lasso}
\end{center}
\end{figure}

The Lasso and SML methods have a comparable computational complexity.  However, unlike the Lasso, the SML method is not parallelizable.  Parallelization would render the Lasso a more computationally attractive choice, since each variable can regressed against all other separately, at an individual cost of $\mathcal{O}(p^3)$.  In exchange, SML can offer a more accurate estimate of the sparsity pattern, as well as a well-conditioned estimate of the covariance matrix itself.

\subsection{Gene expression and U.S. Senate voting records data.}

We tested our algorithms on three sets of data: two gene expression data sets, as well as US Senate voting records.  In this section we briefly explore the resulting graphical models. 

\subsubsection{Rosetta Inpharmatics compendium.}

We applied our algorithms to the Rosetta Inpharmatics Compendium of gene expression profiles described by \cite{Hughes2000}.  The 300 experiment compendium contains $n = 253$ samples with $p = 6136$ variables.  With a view towards obtaining a very sparse graph, we replaced $\alpha/2p^2$ in (\ref{eq:lambda_formula}) by $\alpha$, and set $\alpha = 0.05$.  The resulting penalty parameter is $\lambda = 0.0313$.

This is a large penalty for this data set, and by applying Theorem \ref{thm:independence_property_SML} we find that all but 270 of the variables are estimated to be independent from all the rest, clearly a very conservative estimate.  Figure (\ref{fig:hughesaerial}) displays the resulting graph.

\begin{figure}[ht]
\vskip 0.2in
\begin{center}
\setlength{\epsfxsize}{6in}
\centerline{\epsfbox{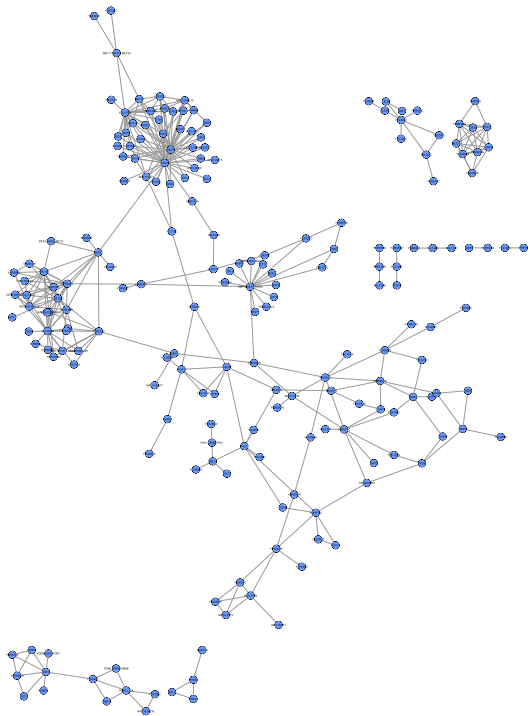}}
\caption{Application to Hughes compendium. The above graph results from solving (\ref{eq:SML}) for this data set with a penalty parameter of $\lambda = 0.0313$.}
\label{fig:hughesaerial}
\end{center}
\end{figure}

Figure (\ref{fig:hughesiron}) closes in on a region of Figure (\ref{fig:hughesaerial}), a cluster of genes that is unconnected to the remaining genes in this estimate.  According to Gene Ontology \citep[see][]{go2000}, these genes are associated with iron homeostasis.  The probability that a gene has been false included in this cluster is at most $0.05$.

\begin{figure}[ht]
\vskip 0.2in
\begin{center}
\setlength{\epsfxsize}{4.5in}
\centerline{\epsfbox{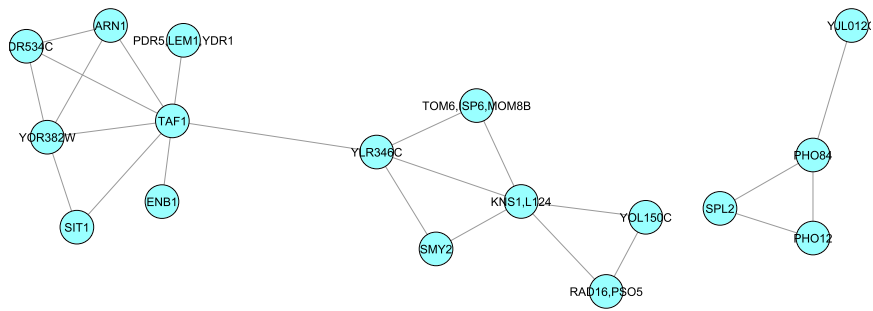}}
\caption{Application to Hughes dataset (closeup of Figure (\ref{fig:hughesaerial}).  These genes are associated with iron homeostasis.}
\label{fig:hughesiron}
\end{center}
\end{figure}

As a second example, in Figure (\ref{fig:hughesfusion}), we show a subgraph of genes associated with cellular membrane fusion.  All three graphs were rendered using Cytoscape.

\begin{figure}[ht]
\vskip 0.2in
\begin{center}
\setlength{\epsfxsize}{4.5in}
\centerline{\epsfbox{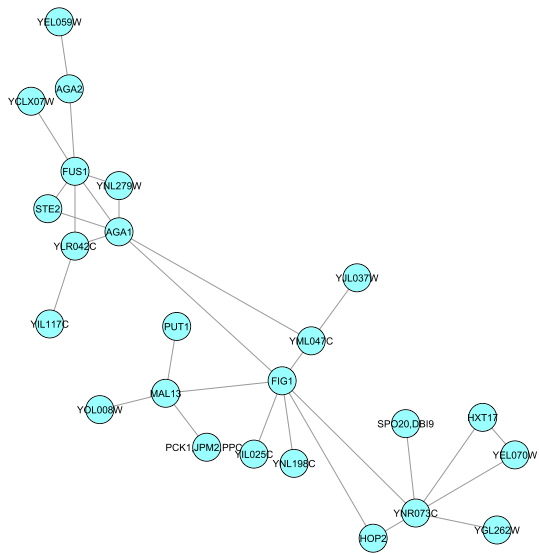}}
\caption{Application to Hughes dataset (subgraph of Figure (\ref{fig:hughesaerial}).  These genes are associated with cellular membrane fusion.}
\label{fig:hughesfusion}
\end{center}
\end{figure}

\subsubsection{Iconix microarray data.}

Next we analyzed a subset of a $10,000$ gene microarray dataset from $160$ drug treated rat livers \cite{natsoulis05}.  In this study, rats were treated with a variety of fibrate, statin, or estrogen receptor agonist compounds.  Taking the 500 genes with the highest variance, we once again replaced $\alpha/2p^2$ in (\ref{eq:lambda_formula}) by $\alpha$, and set $\alpha = 0.05$.  The resulting penalty parameter is $\lambda = 0.0853$. 

By applying Theorem \ref{thm:independence_property_SML} we find that all but 339 of the variables are estimated to be independent from the rest.  This estimate is less conservative than that obtained in the Hughes case since the ratio of samples to variables is 160 to 500 instead of 253 to 6136.

The first order neighbors of any node in a Gaussian graphical model form the set of predictors for that variable.  In the estimated obtained by solving (\ref{eq:SML}), we found that LDL receptor had one of the largest number of first-order neighbors in the Gaussian graphical model.  The LDL receptor is believed to be one of the key mediators of the effect of both statins and estrogenic compounds on LDL cholesterol.  Table \ref{predictorgenes} lists some of the first order neighbors of LDL receptor.

  {\tiny
\begin{table}[tb]
\caption{Predictor genes for LDL receptor.}
\label{predictorgenes}
\vskip 0.15in
\begin{center}
\begin{small}
\begin{sc}
\begin{tabular}{lcccr}
\hline
Accession & Gene \\
\hline
BF553500 & Cbp/p300-interacting transactivator \\
BF387347 & EST \\
BF405996 & calcium channel, voltage dependent \\
NM\underline{ }017158 & cytochrome P450, 2c39 \\
K03249    & enoyl-CoA, hydratase/3-hydroxyacyl Co A dehydrog. \\
BE100965 & EST \\
AI411979  & Carnitine O-acetyltransferase \\
AI410548    & 3-hydroxyisobutyryl-Co A hydrolase \\
NM\underline{ }017288& sodium channel, voltage-gated \\
Y00102      & estrogen receptor 1 \\
NM\underline{ }013200& carnitine palmitoyltransferase 1b\\
\hline
\end{tabular}
\end{sc}
\end{small}
\end{center}
\vskip -0.1in
\end{table}
}

It is perhaps not surprising that several of these genes are directly involved in either lipid or steroid metabolism (K03249, AI411979, AI410548, NM\underline{ }013200, Y00102).  Other genes such as Cbp/p300 are known to be global transcriptional regulators.  Finally, some are un-annotated
ESTs.  Their connection to the LDL receptor in this analysis may provide clues to their function.

\subsubsection{Senate voting records data.}

We conclude our numerical experiments by testing our approximate sparse maximum likelihood estimation method on binary data.  The data set consists of US senate voting records data from the 109th congress (2004 - 2006).  There are one hundred variables, corresponding to 100 senators.  Each of the 542 samples is bill that was put to a vote. The votes are recorded as -1 for no and 1 for yes.

There are many missing values in this dataset, corresponding to missed votes.  Since our analysis depends on data values taken solely from $\{-1, 1\}$, it was necessary to impute values to these.  For this experiment, we replaced all missing votes with noes (-1).  We chose the penalty parameter $\lambda(\alpha)$ according to (\ref{eq:binary_lambda_formula}), using a significance level of $\alpha = 0.05$.  Figure (\ref{fig:allsenators}) shows the resulting graphical model, rendered using Cytoscape.  Red nodes correspond to Republican senators, and blue nodes correspond to Democratic senators. 

\begin{figure}
\begin{center}
\includegraphics[width=5in]{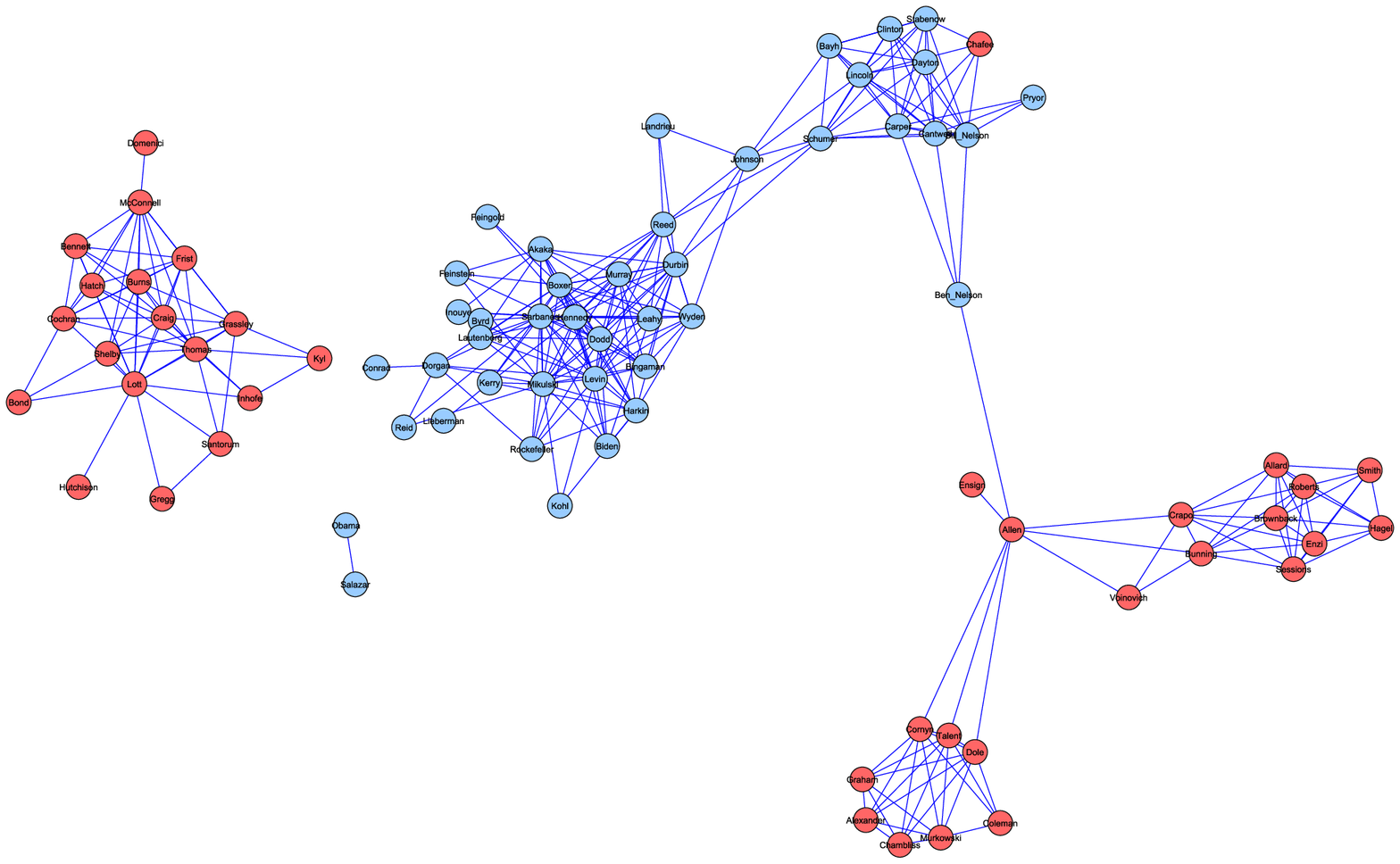}
\end{center}
\caption{US Senate, 109th Congress (2004-2006).  The graph displays the solution to (\ref{eq:binarySML}) obtained using the log determinant relaxation to the log partition function of \cite{Wain2006}.  Democratic senators are colored blue and Republican senators are colored red.}
\label{fig:allsenators}
\end{figure}

We can make some tentative observations by browsing the network of senators.  As neighbors most Democrats have only other Democrats and Republicans have only other Republicans.  Senator Chafee (R, RI) has only democrats as his neighbors, an observation that supports media statements made by and about Chafee during those years.  Senator Allen (R, VA) unites two otherwise separate groups of Republicans and also provides a connection to the large cluster of Democrats through Ben Nelson (D, NE).  Senator Lieberman (D, CT) is connected to other Democrats only through Kerry (D, MA), his running mate in the 2004 presidential election.  These observations also match media statements made by both pundits and politicians.  Thus, although we obtained this graphical model via a relaxation of the log partition function, the resulting picture is largely supported by conventional wisdom.  Figure (\ref{fig:closeupsenators}) shows a subgraph consisting of neighbors of degree three or lower of Senator Allen.  

\begin{figure}
\begin{center}
\includegraphics[width=5in]{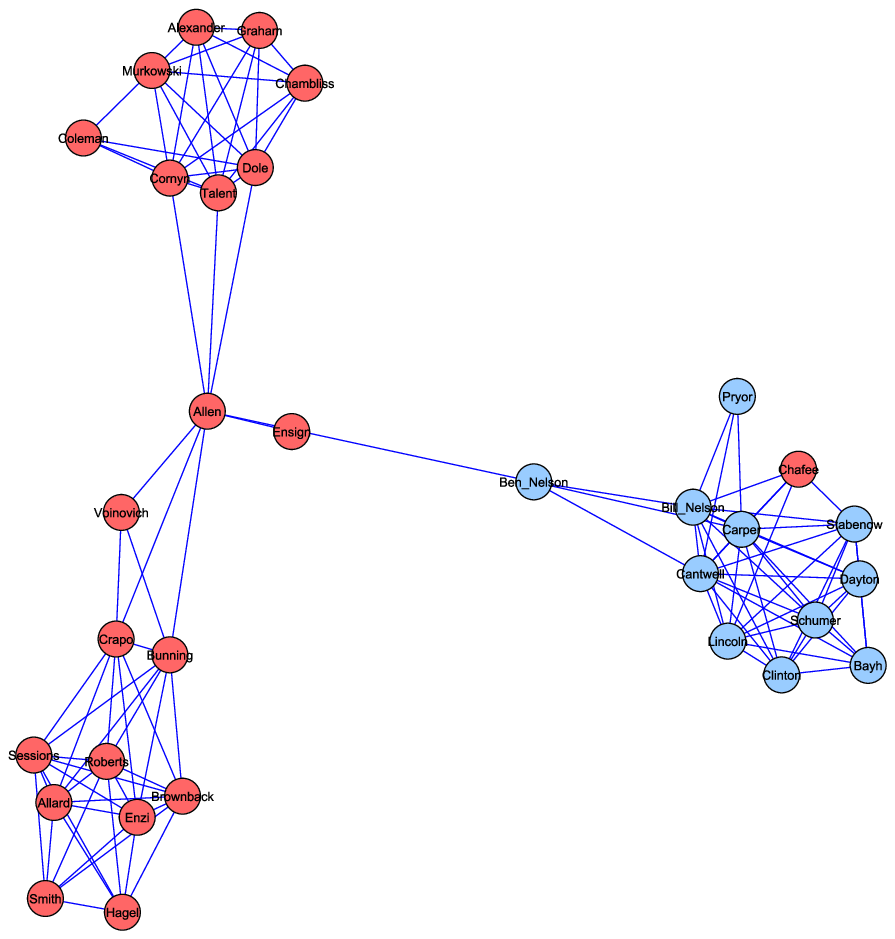}
\end{center}
\caption{US Senate, 109th Congress.  Neighbors of Senator Allen (degree three or lower).}
\label{fig:closeupsenators}
\end{figure}

\acks{We are indebted to Georges Natsoulis for his interpretation of the Iconix dataset analysis and for gathering the senate voting records.  We also thank Martin Wainwright, Bin Yu, Peter Bartlett, and Michael Jordan for many enlightening conversations.}

\bibliography{bibliography}

\begin{thebibliography}{17}
\expandafter\ifx\csname natexlab\endcsname\relax\def\natexlab#1{#1}\fi

\bibitem[Bertsekas(1998)]{bertsekas1998}
D.~Bertsekas.
\newblock {\em Nonlinear Programming}.
\newblock Athena Scientific, 1998.

\bibitem[Consortium(2000)]{go2000}
Gene~Ontology Consortium.
\newblock Gene ontology: tool for the unification of biology.
\newblock {\em Nature Genet.}, 25:\penalty0 25--29, 2000.

\bibitem[Dahl et~al.(2006)Dahl, Roychowdhury, and Vandenberghe]{Dahl05}
J.~Dahl, V.~Roychowdhury, and L.~Vandenberghe.
\newblock Covariance selection for non-chordal graphs via chordal embedding.
\newblock {\em Submitted to Optimization Methods and Software}, 2006.

\bibitem[d'Aspremont et~al.(2004)d'Aspremont, El~Ghaoui, Jordan, and
  Lanckriet]{dasp04a}
Alexandre d'Aspremont, Laurent El~Ghaoui, M.I. Jordan, and G.~R.~G. Lanckriet.
\newblock A direct formulation for sparse {PCA} using semidefinite programming.
\newblock {\em Advances in Neural Information Processing Systems}, 17, 2004.

\bibitem[Dobra and West(2004)]{Dobr04}
A.~Dobra and M.~West.
\newblock Bayesian covariance selection.
\newblock {\em Working paper, ISDS, Duke University}, 2004.

\bibitem[Huang et~al.(2005)Huang, Liu, and Pourahmadi]{Huan05}
J.~Z. Huang, N.~Liu, and M.~Pourahmadi.
\newblock Covariance selection and estimattion via penalized normal likelihood.
\newblock {\em Wharton Preprint}, 2005.

\bibitem[Hughes et~al.(2000)Hughes, Marton, Jones, Roberts, Stoughton, Armour,
  Bennett, Coffey, Dai, He, Kidd, King, Meyer, Slade, Lum, Stepaniants,
  Shoemaker, Gachotte, Chakraburtty, Simon, Bard, and Friend]{Hughes2000}
T.~R. Hughes, M.~J. Marton, A.~R. Jones, C.~J. Roberts, R.~Stoughton, C.~D.
  Armour, H.~A. Bennett, E.~Coffey, H.~Dai, Y.~D. He, M.~J. Kidd, A.~M. King,
  M.~R. Meyer, D.~Slade, P.~Y. Lum, S.~B. Stepaniants, D.~D. Shoemaker,
  D.~Gachotte, K.~Chakraburtty, J.~Simon, M.~Bard, and S.~H. Friend.
\newblock Functional discovery via a compendium of expression profiles.
\newblock {\em Cell}, 102\penalty0 (1):\penalty0 109--126, 2000.

\bibitem[Lauritzen(1996)]{Lauritzen1996}
S.~Lauritzen.
\newblock {\em Graphical Models}.
\newblock Springer Verlag, 1996.

\bibitem[Li and Gui(2005)]{Li2005}
H.~Li and J.~Gui.
\newblock Gradient directed regularization for sparse gaussian concentration
  graphs, with applications to inference of genetic networks.
\newblock {\em University of Pennsylvania Technical Report}, 2005.

\bibitem[Luo and Tseng(1992)]{luo92}
Z.~Q. Luo and P.~Tseng.
\newblock On the convergence of the coordinate descent method for convex
  differentiable minimization.
\newblock {\em Journal of Optimization Theory and Applications}, 72\penalty0
  (1):\penalty0 7--35, 1992.

\bibitem[Meinshausen and B\"uhlmann(2006)]{Mein05}
N.~Meinshausen and P.~B\"uhlmann.
\newblock High dimensional graphs and variable selection with the lasso.
\newblock {\em Annals of statistics}, 34:\penalty0 1436--1462, 2006.

\bibitem[Natsoulis et~al.(2005)Natsoulis, El~Ghaoui, Lanckriet, Tolley, Leroy,
  Dunlea, Eynon, Pearson, Tugendreich, and Jarnagin]{natsoulis05}
G.~Natsoulis, L.~El~Ghaoui, G.~Lanckriet, A.~Tolley, F.~Leroy, S.~Dunlea,
  B.~Eynon, C.~Pearson, S.~Tugendreich, and K.~Jarnagin.
\newblock Classification of a large microarray data set: algorithm comparison
  and analysis of drug signatures.
\newblock {\em Genome Research}, 15:\penalty0 724 --736, 2005.

\bibitem[Nesterov(2005)]{nesterov2003}
Y.~Nesterov.
\newblock Smooth minimization of non-smooth functions.
\newblock {\em Math. Prog.}, 103\penalty0 (1):\penalty0 127--152, 2005.

\bibitem[Tibshirani(1996)]{Tibs96}
R.~Tibshirani.
\newblock Regression shrinkage and selection via the lasso.
\newblock {\em Journal of the {R}oyal statistical society, series {B}},
  58\penalty0 (267-288), 1996.

\bibitem[Vandenberghe et~al.(1998)Vandenberghe, Boyd, and Wu]{vandenberghe98}
Lieven Vandenberghe, Stephen Boyd, and Shao-Po Wu.
\newblock Determinant maximization with linear matrix inequality constraints.
\newblock {\em SIAM Journal on Matrix Analysis and Applications}, 19\penalty0
  (4):\penalty0 499 -- 533, 1998.

\bibitem[Wainwright and Jordan(2006)]{Wain2006}
M.~Wainwright and M.~Jordan.
\newblock Log-determinant relaxation for approximate inference in discrete
  markov random fields.
\newblock {\em IEEE Transactions on Signal Processing}, 2006.

\bibitem[Wainwright et~al.(2006)Wainwright, Ravikumar, and Lafferty]{Wain06_2}
M.~J. Wainwright, P.~Ravikumar, and J.~D. Lafferty.
\newblock High-dimensional graphical model selection using $\ell_1$-regularized
  logistic regression.
\newblock {\em Proceedings of Advances in Neural Information Processing
  Systems}, 2006.

\end{thebibliography}

\appendix
\label{sec:proofs}

{\section{Proof of solution properties and block coordinate descent convergence}
\label{ssec:proof_of_blk_coord_and_properties}}

In this section, we give short proofs of the two theorems on properties of the solution to (\ref{eq:SML}), as well as the convergence of the block coordinate descent method.

{\bf Proof of Theorem \ref{thm:sol_bounds}:}

Since $\hat{\Sigma}$ satisfies $\hat{\Sigma} = S+\hat{U}$, where $\Vert U \Vert_\infty \le \lambda$, we have:
$$
\BA
\Vert \hat{\Sigma} \Vert_2 = \Vert S + \hat{U} \Vert_2 \\
\\
\leq \Vert S \Vert_2 + \Vert U \Vert_2 \leq \Vert S \Vert_2 + \Vert U \Vert_{\infty} \leq \Vert S \Vert_2 + \lambda p \\
\EA
$$
which yields the lower bound on $\Vert \hat{\Sigma}^{-1} \Vert_2$.  Likewise, we can show that $\Vert \hat{\Sigma}^{-1} \Vert_2$ is bounded above.  At the optimum, the primal dual gap is zero:
$$
\BA
-\log \det \hat{\Sigma}^{-1} + \tr(S\hat{\Sigma}^{-1}) + \lambda \Vert \hat{\Sigma}^{-1} \Vert_1 - \log \det \hat{\Sigma} - p\\
\\
= \tr(S\hat{\Sigma}^{-1}) + \lambda \Vert \hat{\Sigma}^{-1} \Vert_1  - p = 0 \\
\EA
$$

We therefore have
$$
\BA
\Vert \hat{\Sigma}^{-1} \Vert_2 \leq \Vert \hat{\Sigma}^{-1} \Vert_F \leq \Vert \hat{\Sigma}^{-1} \Vert_1 \\
\\
= p/\lambda - \tr(S\hat{\Sigma}^{-1})/\lambda \leq p/\lambda \\
\EA
$$
where the last inequality follows from $\tr(S\hat{\Sigma}^{-1}) \geq 0$, since $S \succeq 0$ and $\hat{\Sigma}^{-1} \succ 0$.  {\hfill\BlackBox\\[2mm]}

Next we prove the convergence of block coordinate descent:

{\bf Proof of Theorem \ref{thm:block_coord_descent_convergence}:}

To see that optimizing over one row and column of $W$ in (\ref{eq:SMLdual}) yields the quadratic program (\ref{eq:qpupdate}), let all but the last row and column of $W$ be fixed.  Since we know the diagonal entries of the solution, we can fix the remaining diagonal entry as well:
$$
W = \BP W_{\backslash p \backslash p} & w_p \\ w_p^T & W_{pp} \EP
$$
Then, using Schur complements, we have that 
$$
\det W = \det W_{\backslash p \backslash p} \cdot (W_{pp} - w_p^T(W_{\backslash p \backslash p})^{-1}w_p)
$$
which gives rise to (\ref{eq:qpupdate}).

By general results on block coordinate descent algorithms \citep[e.g.,][]{bertsekas1998}, the algorithms converges if and only if (\ref{eq:qpupdate}) has a unique solution at each iteration.  Thus it suffices to show that, at every sweep, $W^{(j)} \succ 0$ for all columns $j$.  Prior to the first sweep, the initial value of the variable is positive definite: $W^{(0)} \succ 0$ since $W^{(0)} := S + \lambda I$, and we have $S \succeq 0$ and $\lambda > 0$ by assumption.  

Now suppose that $W^{(j)} \succ 0$.  This implies that the following Schur complement is positive:
$$
w_{jj} - W_{j}^T(W_{\backslash j \backslash j}^{(j)})^{-1}W_j > 0
$$
By the update rule we have that the corresponding Schur complement for $W^{(j + 1)}$ is even greater:
$$
w_{jj} - W_{j}^T(W_{\backslash j \backslash j}^{(j+1)})^{-1}W_j > w_{jj} - W_{j}^T(W_{\backslash j \backslash j}^{(j)})^{-1}W_j > 0
$$
so that $W^{(j + 1)} \succ 0$. {\hfill\BlackBox\\[2mm]}

Finally, we apply Theorem \ref{thm:block_coord_descent_convergence} to prove the second property of the solution.

{\bf Proof of Theorem \ref{thm:independence_property_SML}:}

Suppose that column $j$ of the second moment matrix satisfies $\vert S_{ij} \vert \leq \lambda$ for all $i \ne j$.  This means that the zero vector is in the constraint set of (\ref{eq:qpupdate}) for that column.  Each time we return to column $j$, the objective function will be different, but always of the form $y^TAy$ for $A \succ 0$.  Since the constraint set will not change, the solution for column $j$ will always be zero.  By Theorem \ref{thm:block_coord_descent_convergence}, the block coordinate descent algorithm converges to a solution, and so therefore the solution must have $\hat{\Sigma}_j = 0$.  {\hfill\BlackBox\\[2mm]}

{\section{Proof of error bounds.}
\label{ssec:proof_of_error_bounds}}

Next we shall show that the penalty parameter choice given in (\ref{eq:lambda_formula}) yields the error probability bound of Theorem \ref{thm:error_bound}.  The proof is nearly identical to that of \citep[Theorem ~3]{Mein05}.  The differences stem from a different objective function, and the fact that our variable is a matrix of size $p$ rather than a vector of size $p$.  Our proof is only an adaptation of their proof to our problem.

\subsection{Preliminaries}

Before we begin, consider problem (\ref{eq:SML}), for a matrix $S$ of any size:
$$
\hat{X} = \arg\min -\log\det X + \tr(SX) + \lambda\Vert X\Vert_1
$$
where we have dropped the constraint $X \succ 0$ since it is implicit, due to the log determinant function.  Since the problem is unconstrained, the solution $\hat{X}$ must correspond to setting the subgradient of the objective to zero:
\BEQ
\BA
S_{ij} - X^{-1}_{ij} = -\lambda & \hbox{ for } X_{ij} > 0 \\
\\
S_{ij} - X^{-1}_{ij} = \lambda & \hbox{ for } X_{ij} < 0 \\
\\
\vert S_{ij} - X^{-1}_{ij} \vert \leq \lambda & \hbox{ for } X_{ij} = 0 \\
\EA
\label{eq:subgradient_characterization}
\EEQ
Recall that by Theorem \ref{thm:sol_bounds}, the solution is unique for $\lambda$ positive.

\subsection{Proof of error bound for Gaussian data.}

Now we are ready to prove Theorem \ref{thm:error_bound}.  

{\bf Proof of Theorem \ref{thm:error_bound}:}

Sort columns of the covariance matrix so that variables in the same connectivity component are grouped together.  The correct zero pattern for the covariance matrix is then block diagonal.  Define
\BEQ
\Sigma^{\hbox{correct}} := \hbox{blk diag}(C_1, \ldots, C_{\ell})
\EEQ

The inverse $(\Sigma^{\hbox{correct}})^{-1}$ must also be block diagonal, with possible additional zeros inside the blocks.  If we constrain the solution to (\ref{eq:SML}) to have this structure, then by the form of the objective, we can optimize over each block separately.  For each block, the solution is characterized by (\ref{eq:subgradient_characterization}).

Now, suppose that 
\BEQ
\lambda > \max_{i \in N, j \in N \backslash C_i} \vert S_{ij} - \Sigma^{\hbox{correct}}_{ij} \vert
\EEQ
Then, by the subgradient characterization of the solution noted above, and the fact that the solution is unique for $\lambda > 0$, it must be the case that $\hat{\Sigma} = \Sigma^{\hbox{correct}}$.  By the definition of $\Sigma^{\hbox{correct}}$, this implies that, for $\hat{\Sigma}$, we have $\hat{C}_k = C_k$ for all $k \in N$.

Taking the contrapositive of this statement, we can write:
\BEQ
\BA
P(\exists k \in N: \hat{C}_k \not \subseteq C_k) \\
\\
\leq P(\max_{i \in N, j \in N \backslash C_i} \vert S_{ij} - \Sigma^{\hbox{correct}}_{ij} \vert \geq \lambda) \\
\\
\leq p^2(n) \cdot \max_{i \in N, j \in N \backslash C_i} P(\vert S_{ij} - \Sigma^{\hbox{correct}}_{ij} \vert \geq \lambda) \\
\\
=  p^2(n) \cdot \max_{i \in N, j \in N \backslash C_i} P(\vert S_{ij} \vert \geq \lambda) \\
\EA
\EEQ

The equality at the end follows since, by definition, $\Sigma^{\hbox{correct}}_{ij} = 0$ for $j \in N \backslash C_i$. It remains to bound $P(\vert S_{ij} \vert \geq \lambda)$.

The statement $\vert S_{kj} \vert \geq \lambda$ can be written as:
$$
\vert R_{kj} \vert(1 - R_{kj}^2)^{-\frac{1}{2}} \geq \lambda(s_{kk}s_{jj} - \lambda^2)^{-\frac{1}{2}} 
$$  
where $R_{kj}$ is the correlation between variables $k$ and $j$, since
$$
\vert R_{kj} \vert(1 - R_{kj}^2)^{-\frac{1}{2}} =  \vert S_{kj} \vert(S_{kk}S_{jj} - S_{kj}^2)^{-\frac{1}{2}}
$$

Furthermore, the condition $j \in N\backslash C_k$ is equivalent to saying that variables $k$ and $j$ are independent: $\Sigma_{kj} = 0$.  Conditional on this, the statistic 
$$
R_{kj} (1 - R_{kj}^2)^{-\frac{1}{2}}(n - 2)^{\frac{1}{2}}
$$
has a Student's t-distribution for $n - 2$ degrees of freedom.  Therefore, for all $j \in N\backslash C_k$,%
\BEQ
\BA
P(\vert S_{kj} \vert \geq \lambda \vert S_{kk} = s_{kk}, S_{jj} = s_{jj})  \\
\\
= 2P(T_{n - 2} \geq \lambda(s_{kk}s_{jj} - \lambda^2)^{-\frac{1}{2}}(n - 2)^{\frac{1}{2}} \vert S_{kk} = s_{kk}, S_{jj} = s_{jj}) \\
\\
\leq 2\tilde{F}_{n - 2}(\lambda(\hat{\sigma}_k^2\hat{\sigma}_j^2 - \lambda^2)^{-\frac{1}{2}}(n - 2)^{\frac{1}{2}})
\EA
\label{eq:connect_gaussian_binary_lambdas}
\EEQ
where $\hat{\sigma}_k^2$ is the sample variance of variable $k$, and $\tilde{F}_{n-2} = 1 - F_{n-2}$ is the CDF of the Student's t-distribution with $n - 2$ degree of freedom.  This implies that, for all $j \in N\backslash C_k$, 
$$
P(\vert S_{kj} \vert \geq \lambda) \leq 2\tilde{F}_{n - 2}(\lambda(\hat{\sigma}_k^2\hat{\sigma}_j^2 - \lambda^2)^{-\frac{1}{2}}(n - 2)^{\frac{1}{2}})
$$

since $P(A) = \int P(A \vert B)P(B)dB \leq K\int P(B)dB = K$.  Putting the inequalities together, we have that:
$$
\BA
P(\exists k: \hat{C}^{\lambda}_k \not \subseteq C_k) \\
\\
\leq p^2 \cdot \max_{k, j \in N\backslash C_k} 2\tilde{F}_{n - 2}(\lambda(\hat{\sigma}_k^2\hat{\sigma}_j^2 - \lambda^2)^{-\frac{1}{2}}(n - 2)^{\frac{1}{2}}) \\
\\
= 2p^2 \tilde{F}_{n - 2}(\lambda((n - 2)/((\max_{i > j} \hat{\sigma}_k\hat{\sigma}_j)^2 - \lambda^2))^{\frac{1}{2}})
\EA
$$
For any fixed $\alpha$, our required condition on $\lambda$ is therefore
$$
\tilde{F}_{n - 2}(\lambda((n - 2)/((\max_{i > j} \hat{\sigma}_k\hat{\sigma}_j)^2 - \lambda^2))^{\frac{1}{2}}) = \alpha/2p^2
$$
which is satisfied by choosing $\lambda$ according to (\ref{eq:lambda_formula}).{\hfill\BlackBox\\[2mm]}

\subsection{Proof of bound for binary data.}

We can reuse much of the previous proof to derive a corresponding formula for the binary case.

{\bf Proof of Theorem \ref{thm:error_bound_binary}:}

The proof of Theorem \ref{thm:error_bound_binary} is identical to the proof of Theorem \ref{thm:error_bound}, except that we have a different null distribution for $\vert S_{kj}\vert$.  The null distribution of
$$
nR_{kj}^2
$$
is chi-squared with one degree of freedom.  Analogous to (\ref{eq:connect_gaussian_binary_lambdas}), we have:
$$
\BA
P(\vert S_{kj} \vert \geq \lambda \vert S_{kk} = s_{kk}, S_{jj} = s_{jj})  \\
\\
= 2P(nR_{kj}^2 \geq n\lambda^2s_{kk}s_{jj} \vert S_{kk} = s_{kk}, S_{jj} = s_{jj}) \\
\\
\leq 2\tilde{G}(n\lambda^2\hat{\sigma}_k^2\hat{\sigma}_j^2)
\EA
$$
where $\hat{\sigma}_k^2$ is the sample variance of variable $k$, and $\tilde{G} = 1 - G$ is the CDF of the chi-squared distribution with one degree of freedom.  This implies that, for all $j \in N\backslash C_k$, 
$$
P(\vert S_{kj} \vert \geq \lambda) \leq 2\tilde{G}((\lambda\hat{\sigma}_k\hat{\sigma}_j\sqrt{n})^2)
$$

Putting the inequalities together, we have that:
$$
\BA
P(\exists k: \hat{C}^{\lambda}_k \not \subseteq C_k) \\
\\
\leq p^2 \cdot \max_{k, j \in N\backslash C_k} 2\tilde{G}((\lambda\hat{\sigma}_k\hat{\sigma}_j\sqrt{n})^2) \\
\\
= 2p^2 \tilde{G}((\min_{i > j} \hat{\sigma}_k\hat{\sigma}_j)^2n\lambda^2)
\EA
$$
so that, for any fixed $\alpha$, we can achieve our desired bound by choosing $\lambda(\alpha)$ according to (\ref{eq:binary_lambda_formula}).{\hfill\BlackBox\\[2mm]}

{\section{Proof of connection between Gaussian SML and binary ASML}
\label{sec:connect_SML_ASML_proofs}}

We end with a proof of Theorem \ref{thm:connect_ASML_SML}, which connects the exact Gaussian sparse maximum likelihood problem with the approximate sparse maximum likelihood problem obtained by using the log determinant relaxation of \cite{Wain2006}.  First we must prove Lemma \ref{lem:log_partition_bound}.

{\bf Proof of Lemma \ref{lem:log_partition_bound}:}

The conjugate function for the convex normalization $A(\theta)$ is defined as
\BEQ
A^*(\mu) := \sup_{\theta} \{ \langle \mu, \theta \rangle - A(\theta) \}
\EEQ

Wainwright and Jordan derive a lower bound on this conjugate function using an entropy bound:
\BEQ
A^*(\mu) \geq B^*(\mu) 
\EEQ

Since our original variables are spin variables $x ~ \{ -1, +1 \}$, the bound given in the paper is
\BEQ
B^*(\mu) := -\frac{1}{2}\log \det (R(\mu) + \hbox{diag}(m)) - \frac{p}{2}\log(\frac{e\pi}{2} )
\EEQ
where $m := (1, \frac{4}{3}, \ldots, \frac{4}{3})$.

The dual of this lower bound is $B(\theta)$:
\BEQ
\BA
B^*(\mu) := \max_{\theta} \langle \theta, \mu \rangle - B(\theta) \\
\\
\leq \max_{\theta} \langle \theta, \mu \rangle - A(\theta) =: A^*(\mu) \\
\EA
\EEQ

This means that, for all $\mu, \theta$, 
\BEQ
\langle \theta, \mu \rangle - B(\theta) \leq A^*(\mu)
\EEQ
or
\BEQ
B(\theta) \geq \langle \theta, \mu \rangle - A^*(\mu)
\EEQ
so that in particular
\BEQ
B(\theta) \geq \max_{\mu} \langle \theta, \mu \rangle - A^*(\mu) =: A(\theta)
\EEQ
Using the definition of $B(\theta)$ and its dual $B^*(\mu)$, we can write
\BEQ
\BA
B(\theta) := \max_{\mu} \langle \theta, \mu \rangle - B^*(\mu) \\
\\
= \frac{p}{2}\log(\frac{e\pi}{2}) + \max_{\mu} \frac{1}{2}\langle R(\theta), R(\mu) \rangle + \frac{1}{2}\log \det (R(\mu) + \hbox{diag}(m)) \\
\\
= \frac{p}{2}\log(\frac{e\pi}{2}) + \frac{1}{2} \cdot \max \{ \langle R(\theta), X - \hbox{diag}(m) \rangle + \log \det (X) : X \succ 0, \hbox{diag}(X) = m \} \\
\\
= \frac{p}{2}\log(\frac{e\pi}{2}) + \frac{1}{2} \cdot \{ \max_{X \succ 0} \min_{\nu} \langle R(\theta), X - \hbox{diag}(m) \rangle + \log \det (X) + \nu^T(\hbox{diag}(X) - m) \} \\
\\
= \frac{p}{2}\log(\frac{e\pi}{2}) + \frac{1}{2} \cdot \{ \max_{X \succ 0} \min_{\nu} \langle R(\theta) + \hbox{diag}(\nu) , X \rangle + \log \det (X) - \nu^Tm \} \\
\\
= \frac{p}{2}\log(\frac{e\pi}{2}) + \frac{1}{2} \cdot \{ \min_{\nu} - \nu^Tm + \max_{X \succ 0} \langle R(\theta) + \hbox{diag}(\nu) , X \rangle + \log \det (X) \} \\
\\
= \frac{p}{2}\log(\frac{e\pi}{2}) + \frac{1}{2} \cdot \{ \min_{\nu} - \nu^Tm - \log\det (- (R(\theta) + \hbox{diag}(\nu)) ) - (p + 1) \} \\
\\
= \frac{p}{2}\log(\frac{e\pi}{2}) - \frac{1}{2}(p + 1) + \frac{1}{2} \cdot \{ \min_{\nu} - \nu^Tm - \log\det (- (R(\theta) + \hbox{diag}(\nu)) ) \} \\
\\
= \frac{p}{2}\log(\frac{e\pi}{2}) - \frac{1}{2}(p + 1) - \frac{1}{2} \cdot \{ \max_{\nu} \nu^Tm + \log\det (- (R(\theta) + \hbox{diag}(\nu\lambda) \} 
\EA
\EEQ
{\hfill\BlackBox\\[2mm]}

Now we use lemma \ref{lem:log_partition_bound} to prove the main result of section \ref{ssec:binary_problem_formulation}.  Having expressed the upper bound on the log partition function as a constant minus a maximization problem will help when we formulate the sparse approximate maximum likelihood problem.

{\bf Proof of Theorem \ref{thm:connect_ASML_SML}:}

The approximate sparse maximum likelihood problem is obtained by replacing the log partition function $A(\theta)$ with its upper bound $B(\theta)$, as derived in lemma \ref{lem:log_partition_bound}:
\BEQ
\BA
n \cdot \{ \max_{\theta} \frac{1}{2}\langle R(\theta), R(\bar{z}) \rangle - B(\theta) - \lambda \Vert \theta \Vert_1 \} \\
\\
= n \cdot \{ \max_{\theta} \frac{1}{2}\langle R(\theta), R(\bar{z}) \rangle - \lambda \Vert \theta \Vert_1  + \frac{1}{2}(p + 1) - \frac{p}{2}\log(\frac{e\pi}{2}) \\ 
\\
+ \frac{1}{2} \cdot \{ \max_{\nu} \nu^Tm + \log\det (- (R(\theta) + \hbox{diag}(\nu)) )\} \} \\
\\
= \frac{n}{2}(p + 1) - \frac{np}{2}\log(\frac{e\pi}{2}) + \frac{n}{2} \cdot \max_{\theta, \nu} \{ \nu^Tm + \langle R(\theta), R(\bar{z}) \rangle \\
\\
+ \log\det (- (R(\theta) + \hbox{diag}(\nu)) ) - 2\lambda \Vert \theta \Vert_1 \} \\
\EA
\EEQ

We can collect the variables $\theta$ and $\nu$ into an unconstrained symmetric matrix variable $Y := - (R(\theta) + \hbox{diag}(\nu))$.

Observe that 
\BEQ 
\BA
\langle R(\theta), R(\bar{z}) \rangle = \langle - Y - \hbox{diag}(\nu), R(\bar{z}) \rangle \\
\\
= - \langle Y, R(\bar{z}) \rangle - \langle \hbox{diag}(\nu), R(\bar{z}) \rangle = - \langle Y, R(\bar{z}) \rangle \\
\EA
\EEQ
and that
\BEQ
\BA
\nu^Tm = \langle \hbox{diag}(\nu), \hbox{diag}(\nu) \rangle = \langle - Y - R(\theta), \hbox{diag}(m) \rangle \\
\\
= - \langle Y, \hbox{diag}(m) \rangle - \langle R(\theta), \hbox{diag}(m) \rangle = - \langle Y, \hbox{diag}(m) \rangle
\EA
\EEQ

The approximate sparse maximum likelihood problem can then be written in terms of $Y$:
\BEQ
\BA
\frac{n}{2}(p + 1) - \frac{np}{2}\log(\frac{e\pi}{2}) + \frac{n}{2} \cdot \max_{\theta, \nu} \{ \nu^Tm + \langle R(\theta), R(\bar{z}) \rangle \\
\\
+ \log\det (- (R(\theta) + \hbox{diag}(\nu)) ) - 2\lambda \Vert \theta \Vert_1 \} \\
\\
= \frac{n}{2}(p + 1) - \frac{np}{2}\log(\frac{e\pi}{2}) + \frac{n}{2} \cdot \max \{ \log \det Y - \langle Y, R(\bar{z}) + \hbox{diag}(m) \rangle \\ 
\\
- 2\lambda \sum_{i = 2}^p \sum_{j = i + 1}^{p + 1} \vert Y_{ij} \vert \}
\EA
\EEQ

If we let $M := R(\bar{z}) + \hbox{diag}(m)$, then:
$$
M = \BP 1 & \bar{\mu}^T \\ \bar{\mu} & Z + \frac{1}{3}I \EP
\label{eq:binary_sufficientstatistic}
$$
where $\bar{\mu}$ is the sample mean and 
$$
Z = \frac{1}{n}\sum_{k = 1}^n z^{(k)}(z^{(k)})^T
$$
Due to the added $\frac{1}{3}I$ term, we have that $M \succ 0$ for any data set.

The problem can now be written as:
\BEQ
\hat{Y} := \arg \max \{ \log\det Y - \langle Y, M \rangle - 2\lambda \sum_{i = 2}^p \sum_{j = i + 1}^{p + 1} \vert Y_{ij} \vert : Y \succ 0 \}
\label{eq:binary_ASML_primal}
\EEQ

Since we are only penalizing certain elements of the variable $Y$, the solution $\hat{X}$ of the dual problem to (\ref{eq:binary_ASML_primal}) will be of the form:
$$
\hat{X} = \BP 1 & \bar{\mu}^T \\ \bar{\mu} & \tilde{X} \EP
$$
where 
$$
\tilde{X} := \arg \max \{ \log \det V: V_{kk} = Z_{kk} + \frac{1}{3}, \hbox{   } \vert V_{kj} - Z_{kj} \vert \leq \lambda \}.
$$
We can write an equivalent problem for estimating the covariance matrix.  Define a new variable:
$$
\Gamma = V - \bar{\mu}\bar{\mu}^T
$$
Using this variable, and the fact that the second moment matrix about the mean, defined as before, can be written
$$
S = \frac{1}{n}\sum_{k = 1}^n z^{(k)}(z^{(k)})^T - \bar{\mu}\bar{\mu}^T = Z - \bar{\mu}\bar{\mu}^T
$$
we obtain the formulation (\ref{eq:binaryASML_formulation}).  Using Schur complements, we see that our primal variable is of the form:
$$
Y = \BP * & * \\ * & \hat{\Gamma}^{-1} \EP
$$
From our definition of the variable $Y$, we see that the parameters we are estimating, $\hat{\theta}_{kj}$, are the negatives of the off-diagonal elements of $\hat{\Gamma}^{-1}$, which gives us (\ref{eq:binary_ASML_solution}).  {\hfill\BlackBox\\[2mm]}

\end{document}